\documentclass[11pt]{article}

\usepackage[final]{acl}

\usepackage{times}
\usepackage{latexsym}
\usepackage{graphicx}
\usepackage{booktabs}
\usepackage{amsmath}
\usepackage{cleveref}
\usepackage{longtable}
\usepackage{listings}

\usepackage[T1]{fontenc}
\usepackage[utf8]{inputenc}
\usepackage{microtype}
\usepackage{inconsolata}

\graphicspath{{./figure/}}

\hypersetup{
  pdfauthor={Yifei Zhu},
  pdftitle={PolitNuggets: Benchmarking Agentic Discovery of Long-Tail Political Facts}
}

\title{PolitNuggets: Benchmarking Agentic Discovery of Long-Tail Political Facts}

\author{Yifei Zhu \\
  The University of Hong Kong \\
  \texttt{frankyifei@connect.hku.hk} \\}

\begin{document}
\maketitle

\begin{abstract}
  Large Reasoning Models (LRMs) embedded in agentic frameworks have transformed information retrieval from static, long-context question answering into open-ended exploration. Yet real-world use requires models to discover and synthesize “long-tail” facts from dispersed sources, a capability that remains under-evaluated. We introduce \textbf{PolitNuggets}, a multilingual benchmark for agentic information synthesis via constructing political biographies for 400 global elites, covering over 10{,}000 political facts. We standardize evaluation with an optimized Supervisor--Searcher multi-agent system and propose \textbf{FactNet}, an evidence-conditional protocol that scores discovery, fine-grained accuracy, and efficiency. Across models and settings, we find that current systems often struggle with fine-grained details, and vary substantially in efficiency. Finally, using benchmark diagnostics, we relate agent performance to underlying model capabilities, highlighting the importance of short-context extraction, multilingual robustness, and reliable tool use.
\end{abstract}

\section{Introduction}

Reasoning and synthesizing information within a given context is the defining capability of modern Large Reasoning Models (LRMs). The key framework can be called \textbf{Reasoning \emph{in} Context}, where a model is \emph{passively} provided a finite set of evidence and must extract or synthesize answers from it \citep{lewisRetrievalAugmentedGenerationKnowledgeIntensive2020, guuREALMRetrievalAugmentedLanguage2020}. The rapid growth of context windows has enabled strong performance on long-document tasks \citep{shahamZeroSCROLLSZeroShotBenchmark2023, anLEvalInstitutingStandardized2023, zhangInfiniteBenchExtendingLong2024, baiLongBenchV2Deeper2025a, vodrahalliMichelangeloLongContext2024, yangControllableExaminationLongcontext2025, yenHELMETHowEvaluate2024}.

However, a new paradigm is emerging. By integrating LRMs into agentic frameworks equipped with retrieval tools, models can now actively explore, filter, and construct their own context from open-ended sources like the webpages and codebases \citep{nakanoWebGPTBrowserassistedQuestionanswering2021, schickToolformerLanguageModels2023, zhouWebArenaRealisticWeb2023}. This unlocks a different layer of complexity: \textbf{Reasoning \emph{through} Context}. Unlike the passive in-context setting, here the agent must navigate a potentially unbounded information space, making sequential decisions on \emph{what} to read, \emph{when} to stop, and \emph{how} to synthesize fragmented evidence into a coherent whole \citep{weiBrowseCompSimpleChallenging2025}.

\begin{figure}[t]
  \centering
  \includegraphics[width=\columnwidth,height=0.26\textheight,keepaspectratio]{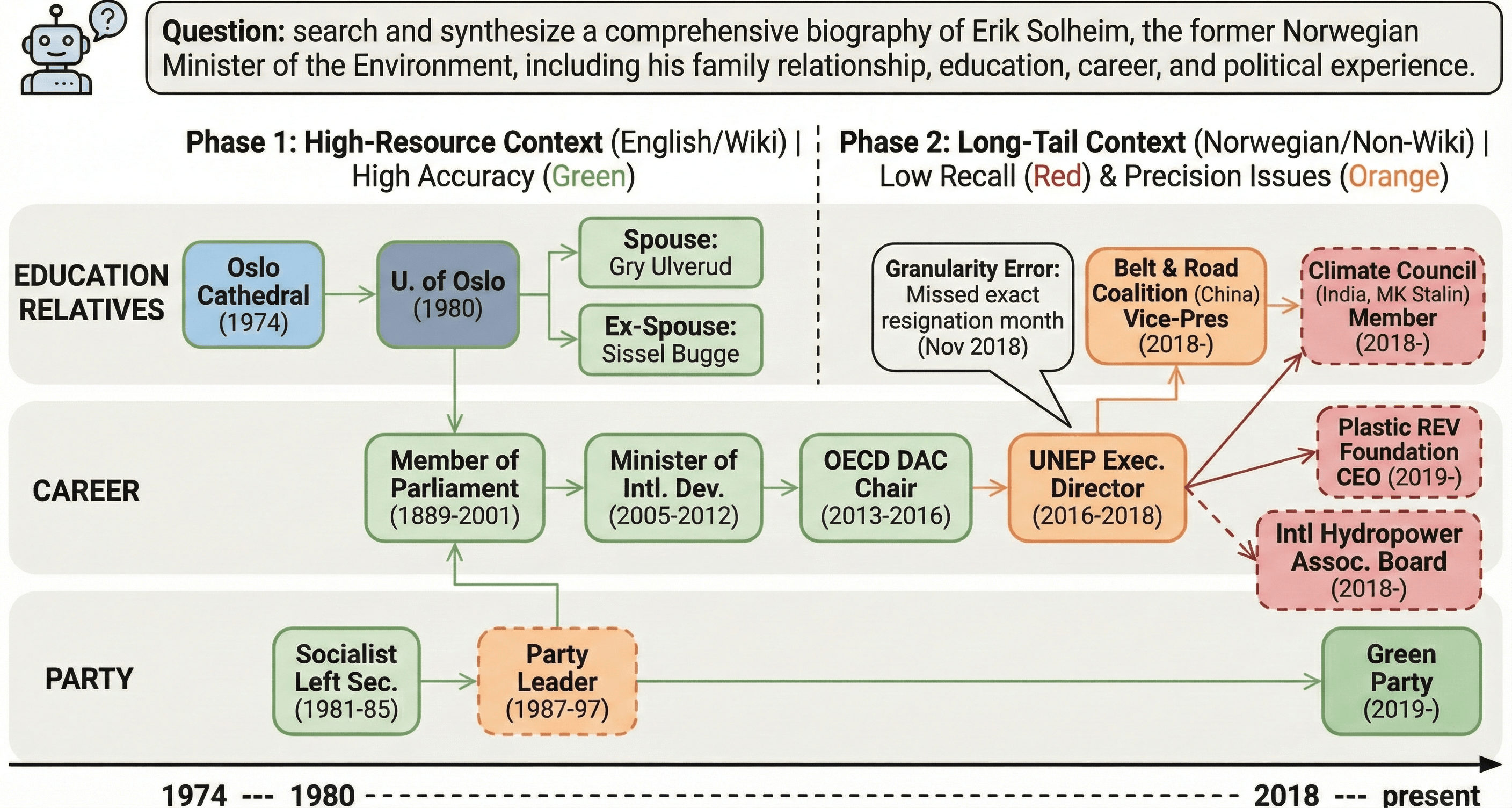}
  \caption{Agent performance heatmap on an example biography (Erik Solheim), illustrating the ``head'' vs.\ ``long-tail'' synthesis gap.}
  \label{fig:teaser}
\end{figure}

While production systems like OpenAI Deep Research \citep{openaiIntroducingDeepResearch2025} and Perplexity Deep Research \citep{perplexityIntroducingPerplexityDeep2025} demonstrate the promise of this agentic paradigm, there remains a lack of rigorous benchmarks for Reasoning \emph{through} Context under longitudinal synthesis demands. Many existing agentic evaluations emphasize short-horizon interactions or isolated fact retrieval \citep{yaoWebShopTowardsScalable2022, mialonGAIABenchmarkGeneral2023, weiBrowseCompSimpleChallenging2025}, and therefore under-measure the professional workflow of reconstructing a coherent narrative from scattered, disconnected, and sometimes contradictory sources. Further, few have linked the performance of a model's reasoning \emph{through} context ability with the performance of a model's reasoning \emph{in} context.

To bridge this gap, we introduce PolitNuggets, a benchmark grounded in a high-impact and realistic task: the construction of political biographies. Wikipedia, while a triumph of collaborative human curation, exhibits systematic coverage gaps---particularly for non-US officials---and often lacks the fine-grained precision required for professional domains like, academic research or political consulting. PolitNuggets tests models' reasoning-through-context abilities by discovering the long-tail biography ``nuggets'' from the open web. This evaluation demands long-horizon reasoning, multi-language understanding, and reliable tool use. Our benchmark also characterizes a static corpus to evaluate models' reasoning-in-context ability. 

Our evaluation of models within an agentic framework reveals that, although agents maintain high precision, they consistently struggle with recall in open-ended settings. We also observe a substantial performance degradation for Non-US entities (up to \(\sim 40\%\) relative drop in F1 in some settings), highlighting a pronounced International Evidence Gap and demonstrating that multilingual robustness is a prerequisite for realistic use. We also connect the reasoning through context ability with the reasoning in context ability. Interestingly, the evaluation results reveal a \textbf{Long-Context Paradox}: strong long-context reading (Reasoning \emph{in} Context) does not reliably predict end-to-end agent performance (Reasoning \emph{through} Context); rather, success is driven by short-context reading precision, reliable tool use, and multi-language understanding.

\subsection{Traversing a latent fact network}
\label{sec:latent_fact_network}
We conceptualize political biography reconstruction not as single-shot retrieval, but as traversing a latent fact network. Let a target biography induce a directed graph \(G=(V,E)\), where nodes \(V\) are atomic ``political nuggets'' (e.g., \emph{Minister of Defense, 2012--2015}) and edges \(E\) are latent temporal/causal links expressed in unstructured text (e.g., \emph{``After resigning in 2015, she joined the World Bank''}). The agent starts from a seed (entity name and minimal metadata) and must recover the relevant subset of \(V\) by expanding along implicit edges discovered in documents.

This induces an optimization trilemma over correctness, coverage, and cost. Agents must maintain high precision (avoid unsupported events), high coverage (high recall over missing events in the long tail), and low efficiency cost (search steps/tokens). This framing explains why naive RAG is insufficient: the missing long-tail nodes may be weakly connected, requiring multi-hop query reformulation and evidence accumulation.

PolitNuggets evaluates whether agents can approximate the full latent fact network while retaining the efficiency of strategic traversal. Strategic traversal jumps between salient nodes (low cost, but vulnerable to missing weakly connected phases if reasoning fails).

\begin{figure}[t]
  \centering
  \includegraphics[width=\linewidth]{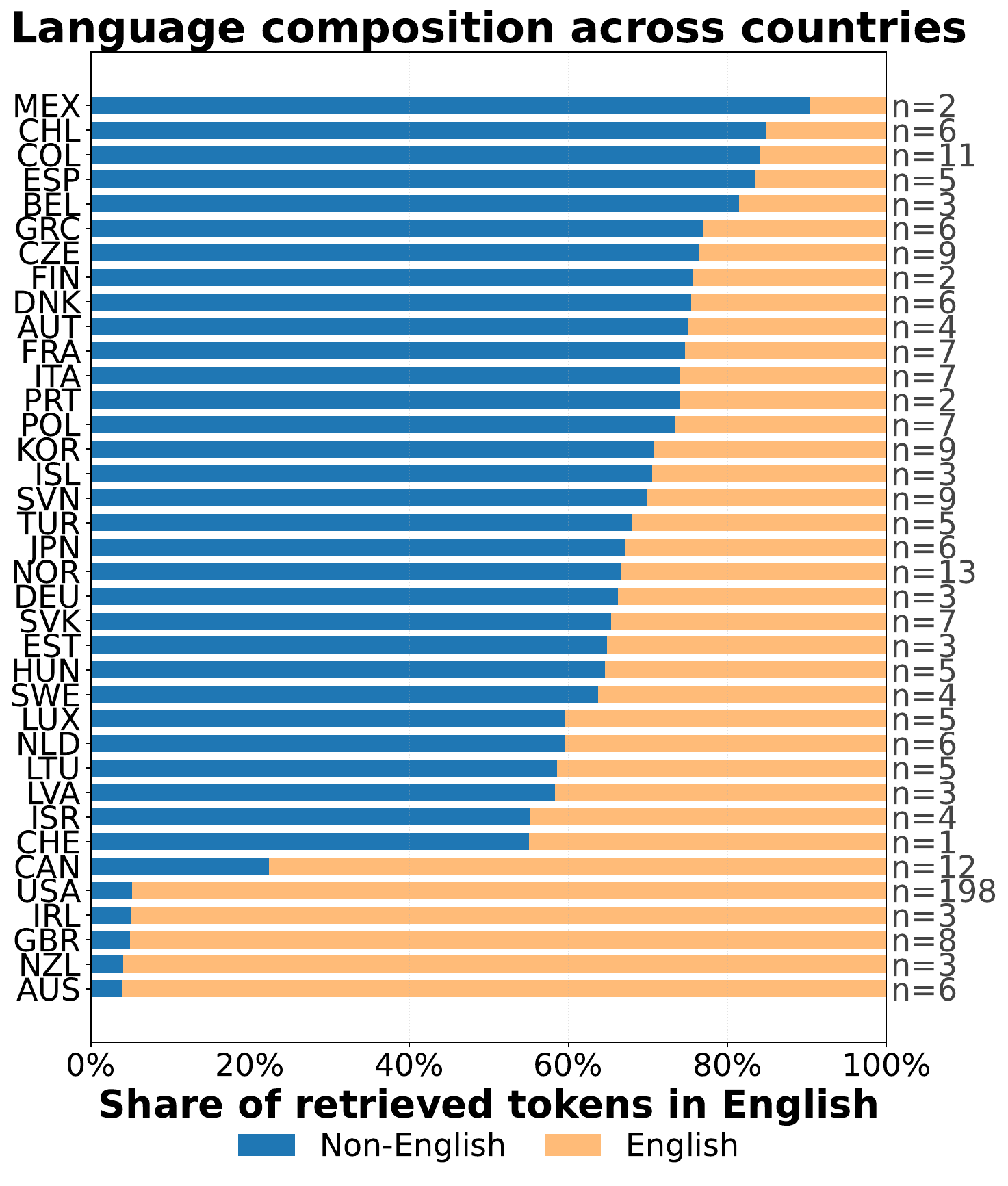}
  \caption{Language composition of retrieved evidence across countries. Bars show the share of retrieved tokens that are English vs.\ non-English; the right-side labels show the number of evaluated cases per country in our benchmark.}
  \label{fig:language_composition}
\end{figure}
\section{Benchmark \& Task}

The PolitNuggets benchmark evaluates agents on their ability to construct accurate, time-resolved career histories for 400 political elites sourced from global government directories.

\subsection{Multilingual evidence in the wild}
The evidence required to reconstruct political careers is inherently multilingual. Traversing a global biography is not merely a search problem: an agent must reason \emph{through} multilingual context to decide what to read next, how to reformulate queries, and when a claim is sufficiently supported. To characterize the language composition of the documents an agent must consume, we compute (for each country) the fraction of retrieved evidence tokens that are in English versus non-English, based on the full set of pages and passages collected during the agentic experiments (Figure~\ref{fig:language_composition}).

Our benchmark instances are drawn from the WhoGov dataset with a US and Non-US sampling design. We randomly sample 200 Non-US cabinet politicians from WhoGov (which provides names and basic metadata for over 58,000 global cabinet members from 1966 to 2023), and we also randomly sample 200 US legislators and senators. After preprocessing and filtering (e.g., ID matching), this yields the 400-entity evaluation set reflected in Figure~\ref{fig:language_composition}.

\subsection{Evaluation Levels: Event-Level vs.\ Attribute-Level}
To disentangle an agent's ability to \emph{find} relevant evidence from its ability to \emph{extract fine-grained details}, we compute F1 at two levels of granularity, following standard slot-filling terminology.

\begin{enumerate}
    \item \textbf{Event-Level F1 (Discovery):} Measures whether the agent correctly identifies the existence of a biographical event. A prediction is a true positive if the Role and Organization match the ground truth and the Year (Start/End) is correct. This primarily tests discovery (did the agent find the right nugget?).
    \item \textbf{Attribute-Level F1 (Granularity):} Measures whether the agent can fill fine-grained attributes for an event (slot filling). A prediction matches only if the event-level criteria are met \emph{and} the Start Month, End Month (within a 1-month tolerance), and Exact Official Title are correct. This primarily tests reading comprehension and schema compliance (did the agent read details correctly?).
\end{enumerate}
The above slot structure (Role/Organization/Date/Title) applies to career and party events; other event types use type-specific key fields (e.g., relation and name for relatives, institution and degree for education) with matching criteria adapted accordingly. Cross-lingual equivalence (e.g., Norwegian titles vs.\ English ground truth) is delegated to the evidence-conditional LLM judge rather than deterministic string normalization.

\subsection{Experiment design and conditions}
\textbf{Model selection.}
To assess the current frontier of agentic information synthesis, we select models that jointly satisfy three constraints required by PolitNuggets: (i) Reasoning \emph{in} Context (strong synthesis from a static context window), (ii) Reasoning \emph{through} Context (robust tool use and multi-turn planning), and (iii) affordability/efficiency (enabling evaluation at the scale of hundreds of entities). As practical proxies, we prioritize models that score highly on OpenAI's Multi-Round Contextual Reasoning (MRCR) benchmark for long-context reasoning \citep{openaiMRCR2025} and on the Berkeley Function Calling Leaderboard (BFCL v3) for tool-use reliability \citep{patilBerkeleyFunctionCalling2025}, while favoring ``Flash/Fast'' variants or efficient open-weight models over prohibitively expensive frontier offerings. This yields our evaluated set: Grok-4-Fast \citep{xaiGrok4FastModel2025}, Gemini-2.5-Flash \citep{geminiteamGemini25Pushing2025}, and Qwen-3 (80B/225B) \citep{qwenteamQwen3TechnicalReport2025}.

\textbf{Task design.}
To disentangle \emph{retrieval} capability from \emph{discovery} capability, we evaluate models in two context conditions:
with Wiki (Enhancement), where the agent is initialized with the target's existing Wikipedia text and must verify claims and fill missing gaps, and without Wiki (Reconstruction), where the agent starts from only the entity's name and must reconstruct the timeline from open-web sources (news archives, government gazettes) under a cold start.

\section{Agentic System}
\subsection{Problem formalization}
\label{sec:problem_formalization}
Let an entity \(e\) have a (latent) biography represented as a set of time-stamped events
\(G_e = \{v_1, \dots, v_n\}\), where each \(v_i = (r_i, o_i, t_i)\) denotes a Role \(r_i\),
Organization \(o_i\), and a time interval \(t_i\) (e.g., start/end year or month).
Let \(W_e \subseteq G_e\) denote the subset covered by the entity's Wikipedia page (when present),
and let \(P_e\) be the set of events predicted by an agent after interacting with the open web.

The agent executes a sequence of search queries \(q_{1:T}\) under a policy \(\pi(q_t \mid h_t)\),
where \(h_t\) is the interaction history (retrieved snippets, intermediate notes, and partial timeline).
Each query incurs a cost \(c(q_t)\) (e.g., a search step and/or token usage), with a budget constraint
\(\sum_{t=1}^T c(q_t) \le C\). The goal is to maximize coverage of missing biography events---i.e.,
high recall on \(G_e \setminus W_e\)---while remaining within budget:
\[
  \max_{\pi}\ \mathrm{E}\big[\mathrm{Recall}(P_e,\ G_e \setminus W_e)\big]
  \quad \text{s.t.}\quad \sum_{t=1}^T c(q_t) \le C.
\]

\subsection{Architecture Details}
\label{sec:architecture_details}

\begin{figure*}[t]
  \centering
  \includegraphics[width=\textwidth]{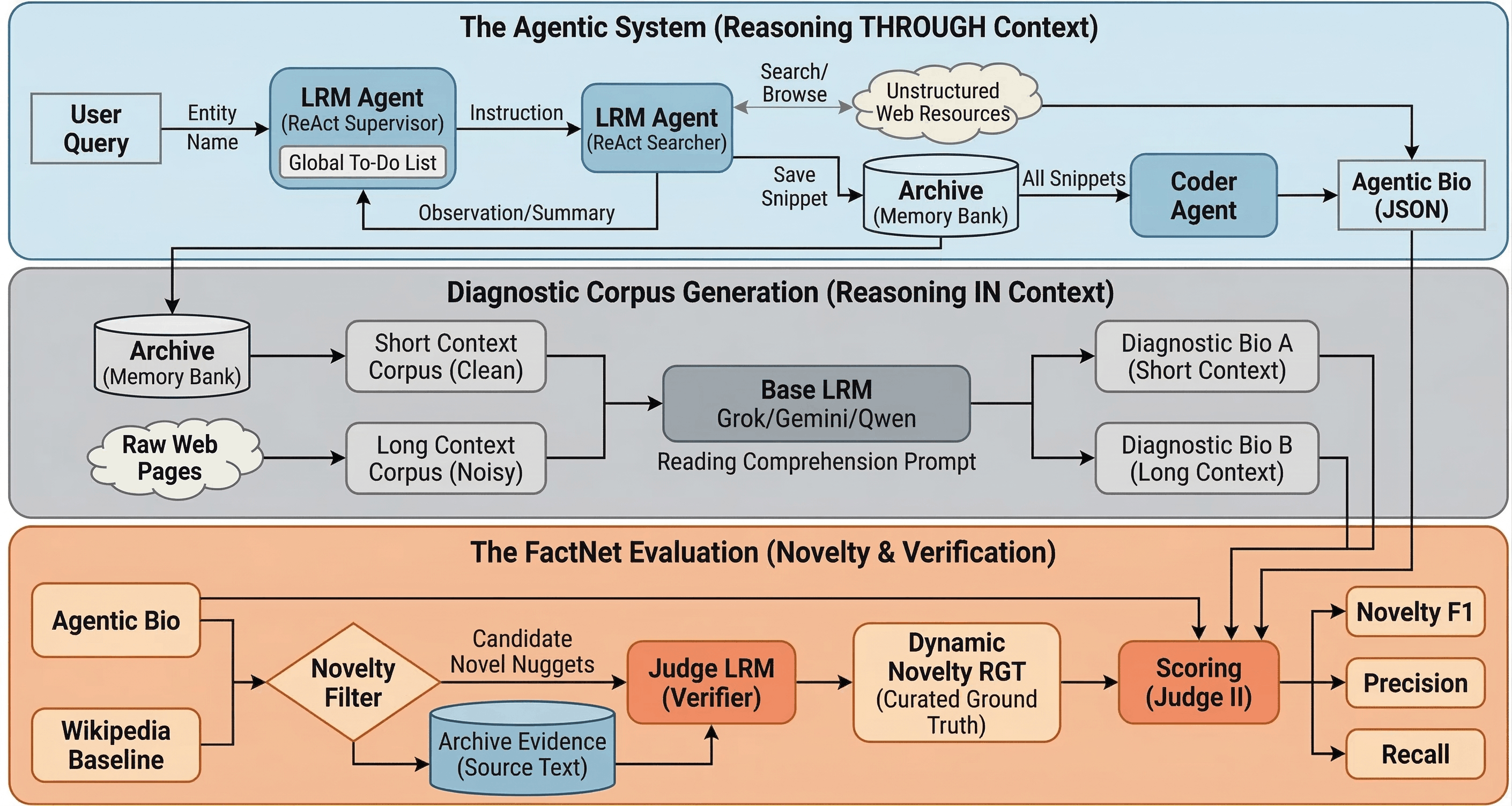}
  \caption{\textbf{The PolitNuggets Framework.} \textbf{(Top) Agentic system:} Supervisor+Searcher (+Archive) produces an Agentic Bio and the evidence corpora (Archive + retrieved pages). \textbf{(Middle) Long-context LRM baselines:} the Base LRM consumes these corpora to produce LRM bios (short-context from Archive; long-context from raw pages). \textbf{(Bottom) FactNet:} evaluates the bios with a dynamic novelty ground truth by filtering Wikipedia-covered facts and validating candidate novel nuggets against archived evidence.}
  \label{fig:system_overview}
\end{figure*}

We implement a standardized Supervisor--Searcher architecture with a clean tool interface to support long-horizon interaction while remaining operationally bounded (Figure~\ref{fig:system_overview}). 
\begin{enumerate}
    \item \textbf{Supervisor:} Maintains global state via a running search summary and a to-do list. It decomposes the biography task into concrete search instructions for the Searcher and decides when to terminate the overall run (e.g., when marginal returns diminish or the step budget is reached).
    \item \textbf{Searcher:} Executes search and browse/retrieve actions over unstructured web resources and returns targeted observations to the Supervisor. In addition to reporting observations, the Searcher can persist related chunks (source-linked evidence snippets) into an Archive. Keeping related records promotes detailed communication. 
\end{enumerate}
Finally, a specialized Coder agent maps the collected evidence into the strict JSON schema required for evaluation. In the final stage, we provide the Coder with both the Supervisor's report (summary + resolved to-do state) and the set of archived related chunks: the report provides global structure and resolved ambiguities, while the raw evidence supplies the fine-grained details needed for attribute filling.

An ablation study shows that adding Archive-backed evidence persistence yields a substantial gain (equivalently, removing the Archive drops Event-Level performance by $\Delta \mathrm{F1} \approx -0.05$), supporting memory as a core design choice (Appendix~\ref{sec:appendix_ablations}).
\paragraph{Architecture vs.\ DeepResearch.}
Empirically, our agentic architecture produces a recall-oriented operating point: the best-performing setting in our system (Grok-4-Fast) achieves higher Event-Level recall than Gemini DeepResearch (powered by gemini 2.5 pro) in the With-wiki condition (US: 0.703 vs.\ 0.678; Non-US: 0.620 vs.\ 0.577), while Gemini DeepResearch is more precision-oriented (EventPrec US: 0.912 vs.\ 0.890; Non-US: 0.892 vs.\ 0.872; Appendix Table~\ref{tab:appendix_prf}).

\section{Evaluation Protocol}

Standard exact-match metrics penalize agents for finding valid information that is absent from the ground truth (false positives). To address this, we employ the FactNet dynamic evaluation protocol. We report F1 at two levels of granularity: Event-Level F1 (discovery of the correct role/organization/year event) and Attribute-Level F1 (strict matching on fine-grained attributes such as start/end month and exact title, conditioned on a correct event).

\subsection{Evaluation design}
\label{sec:eval_design_probes}
Let $G_e$ be the evidence-verified biography nuggets for entity $e$, and let $W_e \subseteq G_e$ denote the subset covered by the entity's Wikipedia page. We construct $G_e$---the \textbf{Consolidated Ground Truth (CGT)}---incrementally from pooled evidence across agentic runs rather than from manual enumeration or Wikipedia. An initial batch of runs produces the seed set; as subsequent runs surface new candidate nuggets, each is verified against the proposing run's archived evidence using the Judge LRM and, if supported, added to $G_e$. All systems are scored against the final snapshot. Wikipedia is used only to define $W_e$ (the coverage filter) and to support the With-Wiki condition. We validate CGT quality via manual timeline inspection (coverage) and human--LLM judge consistency audits plus independent fact-checking via Exa\footnote{\url{https://exa.ai}. Exa is an independent multilingual search backend used for audit.} (precision; Appendix~\ref{sec:appendix_judge_validity}).

Our primary target is the novel set $G = G_e \setminus W_e$ (i.e., facts not already available on Wikipedia at evaluation time). Let $P$ be the set of predicted nuggets produced by a system (agentic bio or LRM bio). We score predictions against a \emph{dynamic novelty ground truth} $G'$, initialized as $G$ and expanded via novelty validation below to avoid penalizing supported discoveries missing from the curated set.

\begin{itemize}
    \item \textbf{Novelty Validation (Dynamic Novelty CGT):} For any predicted nugget $p \in P$ such that $p \notin G$, we treat $p$ as a candidate novel nugget and trigger verification. An external Judge LRM (gpt-5-mini) checks whether $p$ is supported by the system's own evidence (source-linked passages in the Archive). If supported (and not Wikipedia-covered), $p$ is added to $G'$; otherwise it remains a false positive. This yields a Dynamic Novelty CGT that credits verifiable new discoveries while maintaining evidence-grounded precision.
    \item \textbf{Judge reliability checks:} We examined the consistency of this judge with human coders and an external search provider (Exa) via manual re-judging and independent fact checks (Appendix~\ref{sec:appendix_judge_validity}).
    \item \textbf{F1 Score:} Calculated on the dynamic set $G'$:
    \begin{align*}
        \text{Precision} &= \frac{|P \cap G'|}{|P|}, \\
        \text{Recall} &= \frac{|P \cap G'|}{|G'|}, \\
        F_1 &= \frac{2 \cdot \text{Precision} \cdot \text{Recall}}{\text{Precision} + \text{Recall}}.
    \end{align*}
    \item \textbf{Efficiency Cost:} Measured as Average Search Steps per Entity and Total Token Usage. This quantifies the ``cognitive effort'' required to achieve a given F1 score.
\end{itemize}

\subsection{Final evaluation}
\label{sec:lrm_baselines}
We evaluate two families of biographies produced from the same underlying evidence collection runs.
\paragraph{Agentic bios.}
Our agentic system produces Agentic Bios in two context conditions: With Wiki enhancement (4 models: Grok-4-Fast, Gemini-2.5-Flash, Qwen-3-225B, Qwen-3-80B) and Without Wiki reconstruction (2 models: Grok-4-Fast, Gemini-2.5-Flash), yielding 6 agentic bio types in total.
\paragraph{Long-context LRM bios (baselines).}
To quantify ``Reasoning \emph{in} Context'' without agentic search, we ask each Base LRM to generate a biography directly from fixed evidence corpora produced by the Grok-4-Fast With-Wiki runs (the best-performing agentic setting), yielding 8 LRM bio types (4 models $\times$ 2 corpora): (i) a Short-context bio from the curated Archive (fine-grained, deduplicated evidence chunks; \(\sim\)30k tokens on average), and (ii) a Long-context bio from the concatenated Retrieved Web Pages (raw full documents from the same sessions; \(\sim\)300k tokens on average). This isolates improvements attributable to active planning, search, and evidence persistence (Reasoning \emph{through} Context) versus what the Base LRM can achieve from a single static context window.
Importantly, all LRMs are evaluated on the same fixed corpora.

\section{Experimental Results}

\subsection{Main Performance Analysis}

We analyze agent performance through the lens of a three-dimensional evaluation framework: Discovery (Event-Level F1), Granularity (Attribute-Level F1), and Efficiency (search steps/tokens). Table~\ref{tab:main_results} presents the comprehensive performance across all experimental conditions. Grok-4-Fast is the strongest model across both evaluation levels and both contexts, while also using fewer search steps. With Wiki, Grok-4-Fast achieves the best Event-Level F1 (US: 0.768; Non-US: 0.712) and the best Attribute-Level F1 (US: 0.501; Non-US: 0.475) at 11.1 steps on average; without Wiki it remains strong (Event-Level US/Non-US: 0.766/0.708; Attribute-Level US/Non-US: 0.506/0.475) with 14.5 steps. In contrast, Gemini exhibits comparable F1 in some settings but at substantially higher cost in cold-start reconstruction (18.1 steps), and Qwen variants trail in both Event-Level discovery and Attribute-Level slot filling.

\begin{table}[t]
\centering
\footnotesize
\caption{Main results. Performance is reported as F1 at two evaluation levels: Event-Level (discovery of role/organization/year) and Attribute-Level (slot filling of month-level dates and exact titles).}
\label{tab:main_results}
\setlength{\tabcolsep}{4pt}
\begin{tabular}{lllcc}
\toprule
Context & Model & Region & EventF1 & AttrF1 \\
\midrule
With wiki & Gemini DR & US & 0.778 & 0.505 \\
          &        & Non-US & 0.701 & 0.489 \\
With wiki & Gemini & US & 0.638 & 0.407 \\
          &        & Non-US & 0.679 & 0.485 \\
With wiki & grok-4 Fast & US & \textbf{0.768} & \textbf{0.501} \\
          &             & Non-US & \textbf{0.712} & \textbf{0.475} \\
With wiki & qwen-225B & US & 0.499 & 0.335 \\
          &          & Non-US & 0.440 & 0.306 \\
With wiki & qwen-80B & US & 0.510 & 0.344 \\
          &         & Non-US & 0.412 & 0.291 \\
\midrule
Without wiki & Gemini & US & 0.671 & 0.439 \\
             &        & Non-US & 0.618 & 0.468 \\
Without wiki & grok-4 Fast & US & \textbf{0.766} & \textbf{0.506} \\
             &             & Non-US & \textbf{0.708} & \textbf{0.475} \\
\bottomrule
  \end{tabular}
\vspace{0.5em}
\raggedright\scriptsize\emph{Note:} The Without-Wiki condition is limited to Grok-4-Fast and Gemini-2.5-Flash because cold-start reconstruction retrieves substantially more documents, which can exceed Qwen's maximum context window (256k tokens), leading to unstable runs.
\end{table}

\paragraph{Finding 1: Discovery and granularity remain unsolved---and the gap is driven by recall, not precision.}
Performance drops sharply when moving from Event-Level to Attribute-Level evaluation, and even Event-Level discovery is far from saturated. For example, Grok-4-Fast drops from 0.768 to 0.501 F1 (US), showing that extracting month-level dates and exact titles remains difficult. Decomposing performance reveals that this shortfall is primarily a recall/coverage problem rather than a precision problem: precision remains consistently high while recall is substantially lower and deteriorates further at the Attribute-Level (Appendix Table~\ref{tab:appendix_prf}). In other words, agents are largely conservative---they tend to miss weakly connected long-tail events and attributes rather than fabricate unsupported ones.
This precision--recall shape mirrors the passive vs.\ active split: models are often strong at Reasoning \emph{in} Context once a relevant snippet is found, but fail at Reasoning \emph{through} Context when they must autonomously discover that snippet in the first place (Section~\ref{sec:analysis_passive_active} and Table~\ref{tab:appendix_lrm_baseline_results}).

\paragraph{Finding 2: The International Evidence Gap.}
We observe a consistent performance degradation for Non-US entities across most models. While Gemini maintains parity in Event-Level F1 with Wiki (0.638 US vs.\ 0.679 Non-US), Qwen-225B drops from 0.499 (US) to 0.440 (Non-US) at the Event-Level, and performance also degrades at the Attribute-Level across settings. This highlights a structural bias: lower availability of English-language sources and higher complexity in parsing non-US government archives significantly hamper agentic synthesis.

\subsection{Efficiency Analysis: The Pareto Frontier}

\begin{figure}[t]
  \centering
  \includegraphics[width=\linewidth]{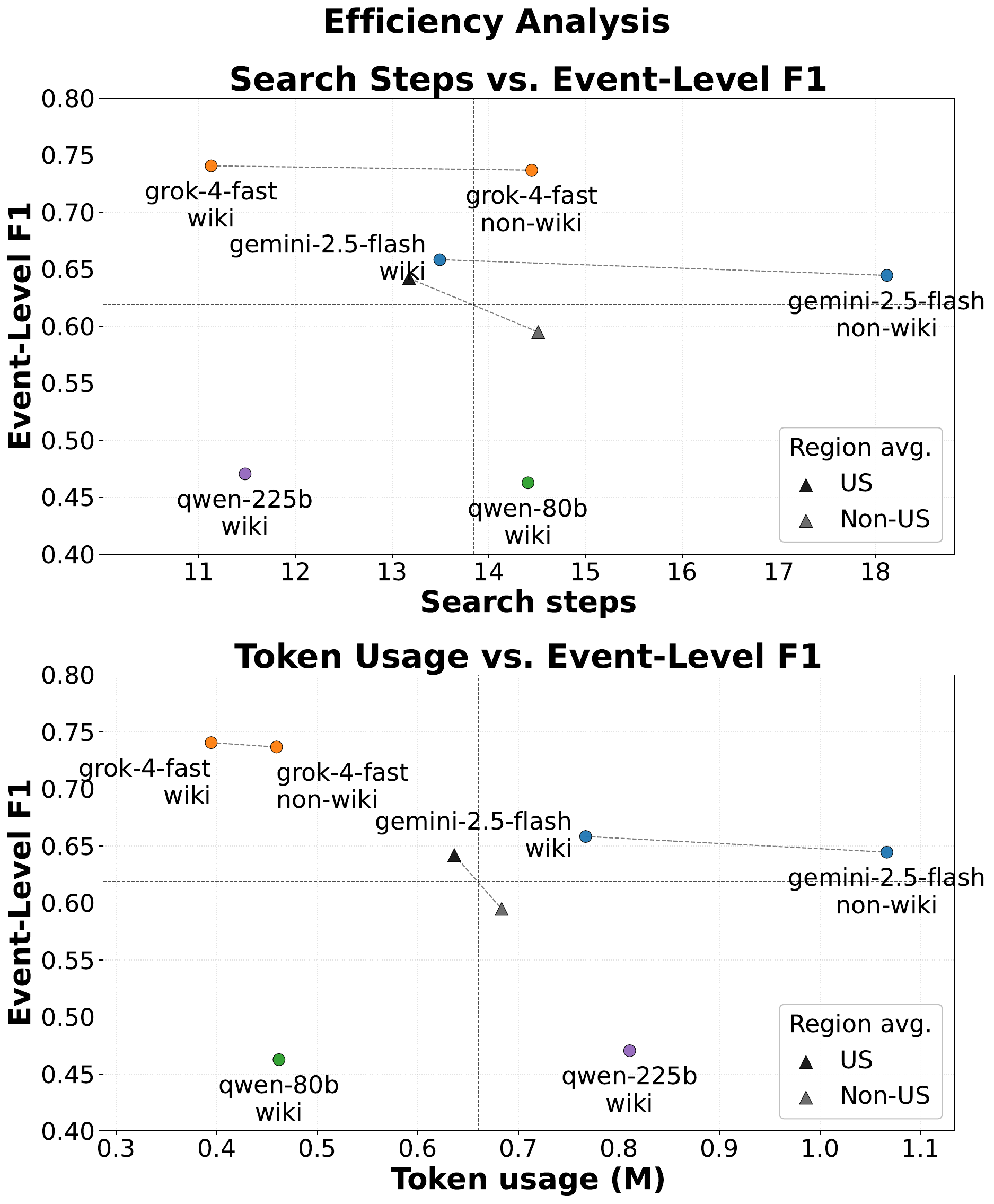}
  \caption{Efficiency Analysis: Search steps vs. F1 (top) and Token usage vs. F1 (bottom). Grok-4-Fast occupies the efficient frontier (top-left), achieving high F1 with minimal steps/tokens---a form of superior cognitive economy. Without Wikipedia context, Gemini maintains similar accuracy but requires substantially higher search volume (long dashed lines), relying on more search rather than improved reasoning efficiency.}
  \label{fig:efficiency}
\end{figure}

To visualize the trade-off between performance and computational cost, we plot the Efficiency Pareto Frontier in Figure~\ref{fig:efficiency}.
The dashed vertical and horizontal reference lines denote the average cost (steps/tokens) and the average F1, splitting the plane into four quadrants. The desirable ``nice zone'' is the top-left quadrant (above-average F1 with below-average cost), while the bottom-right quadrant corresponds to below-average F1 with above-average cost.

\begin{figure*}[t]
  \centering
  \includegraphics[width=\linewidth]{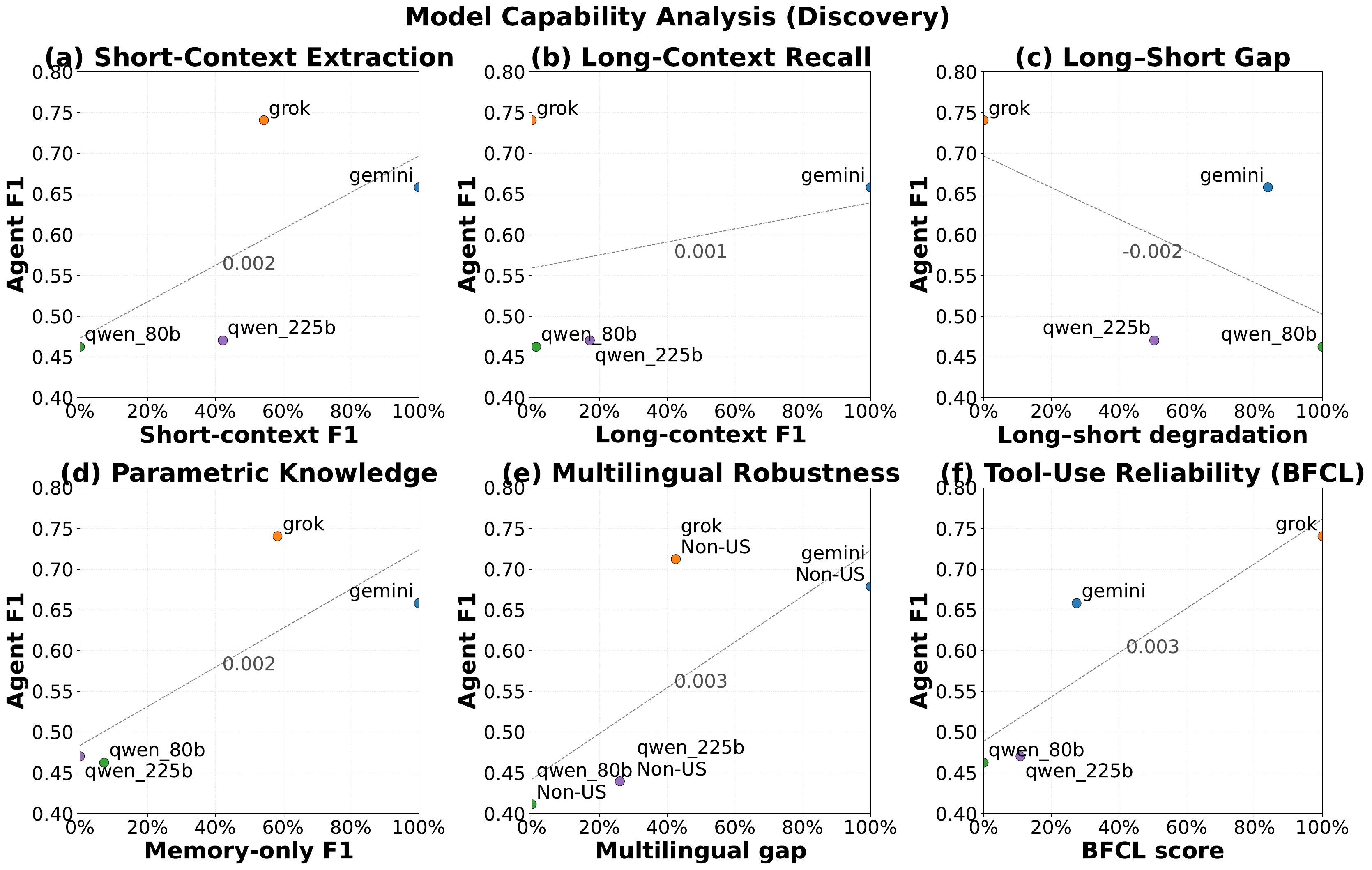}
  \caption{Model capability analysis (Event-Level). Each panel plots a normalized capability score (x-axis) against end-to-end Agent F1 (y-axis). \textbf{(a)} Short-Context Extraction, \textbf{(b)} Long-Context Recall, \textbf{(c)} Long--Short Gap, \textbf{(d)} Parametric Knowledge (closed-book), \textbf{(e)} Multilingual Robustness, \textbf{(f)} Tool-Use Reliability (BFCL). A positive trend indicates that the capability predicts agentic success.}
  \label{fig:model_analysis_easy}
\end{figure*}

Two robustness patterns emerge. First, removing Wikipedia context substantially increases cost (points shift rightward to higher steps/tokens), yet accuracy changes modestly. We interpret this ``Wiki removal'' setting as a stress test of agentic robustness under prolonged trajectories: even as reasoning and search steps rise, F1 does not collapse, suggesting the framework can sustain longer-horizon interaction without severe degradation once it accumulates sufficient evidence. Second, Non-US settings tend to be less efficient than US settings, with many Non-US points shifting toward higher cost and/or lower F1, consistent with a multilingual evidence burden and more fragmented source ecosystems.

\textbf{Finding 3: Wiki removal reveals efficiency gap.} Figure~\ref{fig:efficiency} shows that removing Wiki context consistently shifts points rightward (higher steps/tokens), moving runs out of the top-left ``nice zone.''. Across models, we observe a clear cognitive-economy gap: Grok typically achieves comparable F1 with fewer steps, while Gemini more often substitutes search volume for reasoning precision (a ``brute force'' strategy). Taken together, these results suggest that the core remaining challenge is not simply ``more thinking,'' but more efficient retrieval and search strategy: better query planning and evidence targeting are required to achieve high coverage without paying a large cost increase.
\textbf{Statistical significance.} The key efficiency gaps highlighted here (e.g., Wiki removal increasing steps/tokens for Gemini and Grok) have bootstrap 95\% confidence intervals for mean deltas that exclude 0; see Appendix~\ref{sec:appendix_statistics}.

\section{LRM Analysis: Bridging Passive Reasoning and Active Discovery}
\label{sec:analysis_passive_active}

Having established the agentic performance benchmarks in Section~\ref{tab:main_results} and the controlled ``Reasoning \emph{in} Context'' baselines in Section~\ref{sec:lrm_baselines}, we now investigate the fundamental drivers of success in longitudinal information synthesis. Using the short/long-context corpora defined in Section~\ref{sec:lrm_baselines} and the FactNet protocol in Section~\ref{sec:eval_design_probes}, we decompose performance into six diagnostic dimensions: Short-Context Extraction, Long-Context Recall, the Long--Short Gap, Parametric Knowledge, Multilingual Robustness, and Tool-Use Reliability (BFCL). This analysis aims to bridge the gap between traditional ``Reasoning \emph{in} Context'' benchmarks and the emerging ``Reasoning \emph{through} Context'' paradigm. For completeness, we report the full LRM baseline results table in Appendix~\ref{sec:appendix_full_experiment_results} (Table~\ref{tab:appendix_lrm_baseline_results}).

\subsection{The primacy of short-context extraction}
We observe that a model's ability to extract facts from a clean, short context---the curated Archive---is strongly predictive of end-to-end agentic performance (Figure~\ref{fig:model_analysis_easy}a). This suggests a ``last-mile'' bottleneck: if the model cannot reliably parse and structure high-quality evidence it has already found, additional searching cannot recover the loss, and end-to-end synthesis degrades primarily through missed events and attributes.

\subsection{The decoupling of long-context recall}
Crucially, we find that passive long-context recall on the noisy, full-document baseline is a weak predictor of agentic success. In particular, models that dominate end-to-end discovery do not necessarily outperform peers on static long-context extraction. This validates an \emph{agentic hypothesis}: episodic search that iteratively curates small, high-quality contexts can outperform reliance on massive context windows alone, effectively bypassing the limitations of ``Reasoning \emph{in} Context'' under noise.

An unintuitive finding is that the Long--Short gap is not a reliable proxy for agentic success. Degradation from curated short contexts (Archive) to noisy long contexts (full retrieved pages) does not consistently track end-to-end agent F1 (Figure~\ref{fig:model_analysis_easy}b--c). One plausible explanation is a \emph{training dilemma}. If so, ``better long-context'' and ``smaller long--short degradation'' may not co-exist monotonically under realistic budget and model-structure constraints.

\subsection{The multilingual reasoning barrier}
The ``International Evidence Gap'' observed in our main results is structurally explained by a multilingual reasoning barrier (Figure~\ref{fig:model_analysis_easy}e). Models that exhibit larger degradation when extracting from non-English evidence chunks also underperform on Non-US entities. This indicates that global longitudinal synthesis is bottlenecked not only by retrieving foreign-language documents, but by reasoning over them with comparable fidelity to English evidence.

\subsection{Parametric knowledge and tool reliability as scaffold}
Finally, we observe a positive relationship between a model's closed-book (no-evidence) biography ability and its capacity to discover missing facts (Figure~\ref{fig:model_analysis_easy}d,f). Parametric knowledge appears to act as a semantic scaffold: it supports entity disambiguation, improves query formulation, and helps the agent recognize valuable nuggets when encountered in the wild. Importantly, this semantic scaffold only helps end-to-end if the model can \emph{reliably act on it}: tool-use reliability (BFCL) complements parametric knowledge by reducing brittle failures in search/browse execution, enabling consistent multi-step query reformulation, and translating high-level intent into stable tool calls.

\section{Related Works}
\label{sec:related_work}

\textbf{Evaluating reasoning \emph{in} context:} A rich line of long-context benchmarks has progressively raised the difficulty of passive evidence extraction. Early ``needle-in-a-haystack'' setups (e.g., HELMET \citep{yenHELMETHowEvaluate2024}) probe whether models can locate a single target fact in a long context. Subsequent benchmarks extend this to multiple needles and multi-round contextual reasoning (MRCR \citep{openaiMRCR2025}), and further to structured reasoning over latent or explicit graphs (Michelangelo \citep{vodrahalliMichelangeloLongContext2024}, GraphWalks \citep{openaiGraphWalks2025}). Closest to our setting, LongBioBench \citep{yangControllableExaminationLongcontext2025} uses controlled synthetic biographies to examine long-context understanding, reasoning, and trustworthy generation. PolitNuggets progresses from this lineage by shifting the locus of difficulty from reasoning \emph{in} a given context to reasoning \emph{through} context---where the agent must actively discover, filter, and synthesize evidence from the open web---and by grounding evaluation in real-world, multilingual political biographies rather than synthetic text.

\textbf{Evaluating reasoning \emph{through} context: } Evaluation of tool-augmented reasoning spans static multi-hop QA datasets such as MuSiQue \citep{trivediMuSiQueMultihopQuestions2022} and general agentic benchmarks such as GAIA \citep{mialonGAIABenchmarkGeneral2023} that assess fundamental tool proficiency. More recently, benchmarks including WebSailor \citep{liWebSailorNavigatingSuperhuman2025} and BrowseComp \citep{weiBrowseCompSimpleChallenging2025} emphasize verifying retrieved information in open environments, often following a ``hard-to-find, easy-to-verify'' paradigm (e.g., identifying a specific paper from indirect cues) \citep{weiBrowseCompSimpleChallenging2025}. DeepResearch Bench \citep{duDeepResearchBenchComprehensive2025} pushes toward realistic research workflows with expert-crafted tasks, but this style of evaluation can be expensive and remains sensitive to verification quality. PolitNuggets builds on this line while moving beyond isolated fact lookup to multi-faceted biography synthesis, and provides a scalable evaluation protocol for multilingual discovery, addressing a critical gap in global agentic information retrieval.

\section{Conclusion}

PolitNuggets provides a rigorous assessment of agentic information synthesis in the wild, targeting the gap between Reasoning \emph{in} Context (fixed evidence) and Reasoning \emph{through} Context (tool-driven discovery). We introduce a scalable benchmark of political biography construction for 400 global elites and evaluate systems with FactNet, an evidence-conditional protocol that measures discovery, fine-grained accuracy, and efficiency while validating candidate nuggets against retrieved evidence. Across models and settings, we find that precision is generally high but performance is recall- and efficiency-limited, with Wikipedia removal substantially increasing search cost, and we observe a pronounced International Evidence Gap on Non-US entities. Ablations show that evidence persistence improves end-to-end outcomes, and our diagnostics highlight a ``Long-Context Paradox'': strong long-context reading does not reliably predict agentic success, which is instead driven by short-context extraction, multilingual robustness, and reliable tool use. We hope PolitNuggets and its released artifacts support reproducible progress on evaluating and improving real-world agentic systems.

\section*{Acknowledgments}
We thank Songpo Yang, Xiao Liu, Jiangnan Zhu, and Junyan Jiang for insightful discussions around this project. We also thank the anonymous reviewers for their constructive feedback.

\bibliography{custom}

\section{Limitations}
\label{sec:limitations}
First, due to budget constraints and practical model selection, we do not evaluate the largest and most expensive frontier-scale models. Such models may reveal a clearer connection (or a different relationship) between Reasoning \emph{in} Context and Reasoning \emph{through} Context.

Second, although we provide cached results for reproducibility, benchmark outcomes may still shift over time due to changes in the underlying search engine and the evolving web (ranking drift, content updates, and availability).

Third, our static-context LRM baselines are constructed from evidence corpora produced by agentic collection runs. This yields a controlled comparison, but it may not fully represent long-context performance under independently collected evidence. Thus, we can not conclude that reasoning through context is better than reasoning in context in this research.

\section{Ethical Considerations}
\label{sec:ethics}
PolitNuggets is constructed from human-related information available in the public domain (e.g., Wikipedia, official government pages, and public news/biographical sources). We adhere to fair-use principles for research and release only cached materials necessary for replication. We do not intentionally collect or disclose private or sensitive personal information beyond what is already publicly available, and we do not include any leaked/private datasets.

\section{Potential Risks}
\label{sec:risks}
The agentic biography-construction techniques evaluated here could in principle be repurposed to profile private individuals or non-public figures; we therefore restrict our benchmark to public officials whose career information is a matter of public record. Model-generated biographies can also contain factual errors that, if redistributed uncritically, could harm the subjects or propagate misinformation; we mitigate this by scoring only against evidence-verified nuggets and by releasing the cached source pages so that each claim can be audited. Finally, the International Evidence Gap we document indicates that naive deployment of such systems may disproportionately under-represent non-US and non-English political figures, and this limitation should be explicitly disclosed in any downstream use.

\section{AI Usage}
\label{sec:ai_usage}
We used AI tools to assist with (i) implementation and refactoring of the evaluation and analysis code, (ii) plotting and figure formatting, and (iii) proofreading and minor style edits of the manuscript. All experimental design decisions, evaluations, and reported results were reviewed by the authors, and we validated code changes by re-running the pipeline to reproduce tables and figures.

\appendix

\section{Additional Analysis}
\label{sec:appendix}

\subsection{Experiment Details}
\label{sec:appendix_experiment_details}

\subsubsection{Deliverables}
\label{sec:appendix_deliverables}
To support replication, we release a code repository\footnote{\url{https://github.com/yifeifrank/poli_searcher}} containing the Supervisor--Searcher agentic pipeline, together with a data release\footnote{\url{https://huggingface.co/datasets/frankyifei/politnuggets}} containing the following three artifacts:
\begin{enumerate}
    \item \textbf{Consolidated Ground Truth (CGT).} The final pooled, evidence-verified biography nuggets for all 400 entities (including the Wikipedia-coverage filter $W_e$), which define the evaluation target $G$ and the dynamic novelty set $G'$.
    \item \textbf{Cached webpages.} The raw retrieved web pages collected during our agentic runs, fixing the search snapshot used for all reported numbers and enabling offline re-evaluation.
    \item \textbf{LRM evaluation package.} A curated static-context dataset (Archive-style short context and long-context corpora derived from the cached pages) for evaluating long-context biography extraction without interactive search, enabling controlled comparison of ``Reasoning \emph{in} Context'' across models.
\end{enumerate}
All LRM-baseline and FactNet evaluation procedures are fully specified by the prompts in Appendix~\ref{sec:appendix_prompts} together with the released artifacts above, allowing reassembly without a dedicated code release.

\paragraph{Infrastructure, tools, and cost accounting.}
We implemented the full Supervisor--Searcher pipeline in \texttt{langgraph}.
For LLM inference, we used OpenRouter as the API provider and recorded token usage using OpenRouter's standardized token accounting.
For web search, we used the Serper API; for page retrieval, we used the scrawling service by Jina and Exa. To maximize robustness at scale, we used multiple layers of retry/backoff to ensure successful search and retrieval.
Across the full experimental campaign (including development/testing), we issued approximately \(300\text{k}\) searches (about \$300 in search cost) and retrieved pages at an additional cost of about \$50 (including free-tier, in-limit usage).
Overall, the total third-party API spend for the project (including LLM APIs, search, and retrieval) was approximately \$3{,}750.

\paragraph{Budget controls and termination criteria.}
We enforced two complementary termination criteria to bound the system's budget. First, the Supervisor maintains a to-do list and allocates a bounded amount of \emph{dedicated research} per item: for each to-do item, the Searcher is allowed at most three focused search--retrieve attempts; if the item remains unresolved after three attempts, the Supervisor abandons that branch and proceeds to other items to avoid pathological loops. Second, we impose a hard-coded global cap of 100 LLM calls per run to bound worst-case cost and latency.

\paragraph{Experimental timeline and operational exclusion.}
Our primary data collection was conducted in September 2025 (Gemini and Qwen families), and we added Grok-4-Fast in November 2025 given its strong tool-use reliability and direct relevance to real-world agentic deployments. We also attempted to evaluate GPT-5-Mini, but in preliminary runs it frequently failed to terminate within the maximum conversation times budget ($T=100$), exhibiting repetitive search loops and unfinished trajectories; we therefore exclude it from the final comparison to preserve integrity of the evaluation. However, due to it's strong long context performance and affordability, we used as the judge LRMs in this article.

\paragraph{Consistency and validity of the LLM judge.}
\label{sec:appendix_judge_validity}
To assess the reliability of our evidence-conditional LLM judge, we manually re-judged the final scores for 40 randomly selected officials and compared them against the LLM judge outputs. The human and LLM judgments are consistent, with a correlation of 0.87.
As an additional validity check, we used Exa to further fact-check a random sample of 100 officials: out of 2{,}243 validated nuggets, we identified 82 false positives, corresponding to an inaccuracy rate of \(\approx 3.66\%\).
The manual re-judging was performed by four student annotators recruited through the authors' university network. Each annotator was compensated at HKD 70 per hour, which is the prevailing rate for comparable student research assistance in Hong Kong. Annotators were informed in writing about the purpose of the task (validating LLM-generated biography assessments of public political figures), the use of their labels (aggregate statistics only, no personal data collected). The instruction given are similar to prompts used in research.

\begin{figure}[t]
  \centering
  \includegraphics[width=\linewidth]{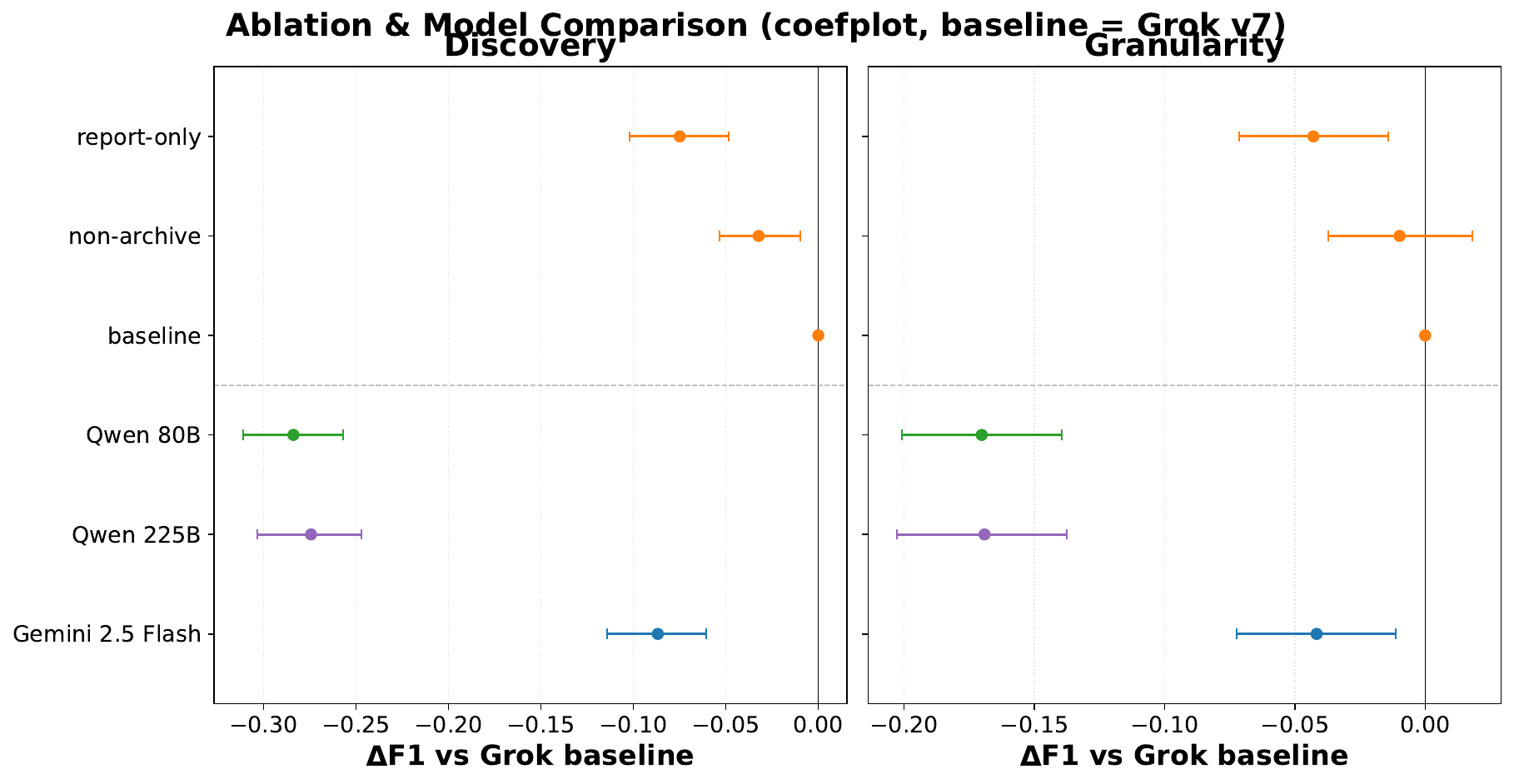}
  \caption{Architectural Ablation (Coefplot). Removing the Archive (``non-archive'') significantly degrades performance ($\Delta \text{F1} \approx -0.05$), confirming that raw evidence persistence is crucial for longitudinal synthesis.}
  \label{fig:ablations}
\end{figure}

\paragraph{Architectural ablation: the necessity of memory.}
\label{sec:appendix_ablations}
To validate the Archive memory mechanism added on top of the base Supervisor--Searcher loop, we conduct ablation studies on the Grok-4 baseline (Figure~\ref{fig:ablations}). We compare the full system against a No-Archive variant (where evidence persistence is disabled and the Supervisor relies only on the Searcher's summaries) and a Report-Only variant.

The results are unequivocal. The coefficient for non-archive is significantly negative ($\beta \approx -0.05$ at the Event-Level), validating our hypothesis: summaries are lossy. Without the Archive Tool to persist raw evidence chunks, the agent suffers from ``Contextual Amnesia,'' hallucinating connections or failing to link dependent facts.

\subsection{Statistical significance}
\label{sec:appendix_statistics}
To support the claims about setting and region gaps, we report distributional summaries and bootstrap confidence intervals for the underlying per-entity metrics. Table~\ref{tab:appendix_stats_diffs} provides key contrasts (mean deltas with bootstrap 95\% CIs; ``CI excludes 0'' indicates statistical significance at the 5\% level).

In table 2, each row corresponds to one \emph{difference} computed over entities.
\textbf{comparison} specifies the direction:
(\texttt{Non-US minus US}) is $\mathrm{E}[\text{Non-US}] - \mathrm{E}[\text{US}]$ within a fixed context/model;
(\texttt{Without wiki minus With wiki}) is $\mathrm{E}[\text{Without}] - \mathrm{E}[\text{With}]$ for the same model.
\textbf{level} indicates whether the outcome is \textbf{Discovery} (event-level), \textbf{Granularity} (attribute-level), or \textbf{Efficiency} (cost).
\textbf{metric} is the quantity being compared (\texttt{f1}, \texttt{steps}, or \texttt{tokens\_total}).
\textbf{n\_us} and \textbf{n\_non\_us} are the effective sample sizes (completed entities) for the two groups, and \textbf{mean\_us} / \textbf{mean\_non\_us} are the corresponding sample means.
\textbf{delta} is the mean difference in the direction defined by \textbf{comparison}.
\textbf{ci95\_low} and \textbf{ci95\_high} are the bootstrap 95\% confidence interval bounds for \textbf{delta}, and \textbf{ci95\_excludes\_0} is \texttt{True} when the CI does not cross 0 (significant at $\alpha=0.05$).
\textbf{cohen\_d} reports a standardized effect size (sign follows \textbf{delta}).
We use unpaired bootstrap resampling for \texttt{Non-US minus US} (different entity sets) and paired bootstrap for \texttt{Without wiki minus With wiki} when comparing the same entities across settings.

\begin{table*}[t]
\centering
\scriptsize
\setlength{\tabcolsep}{3pt}
\caption{Statistical evidence for key differences. We report mean deltas with bootstrap 95\% confidence intervals (CIs) computed over entities; ``CI excludes 0'' indicates significance.}
\label{tab:appendix_stats_diffs}
\resizebox{\textwidth}{!}{\begin{tabular}{lllllrrrrrrrrr}
\toprule
comparison & level & context & model & metric & n\_us & n\_non\_us & mean\_us & mean\_non\_us & delta & ci95\_low & ci95\_high & ci95\_excludes\_0 & cohen\_d \\
\midrule
Non-US minus US & Discovery & With wiki & grok-4 Fast & f1 & 193 & 189 & 0.7682 & 0.7125 & -0.0557 & -0.0958 & -0.0163 & True & -0.2852 \\
Non-US minus US & Granularity & With wiki & grok-4 Fast & f1 & 193 & 189 & 0.5011 & 0.4755 & -0.0256 & -0.0725 & 0.0213 & False & -0.1105 \\
Non-US minus US & Discovery & Without wiki & grok-4 Fast & f1 & 185 & 186 & 0.7656 & 0.7081 & -0.0575 & -0.0948 & -0.0188 & True & -0.3040 \\
Non-US minus US & Granularity & Without wiki & grok-4 Fast & f1 & 185 & 186 & 0.5056 & 0.4748 & -0.0307 & -0.0724 & 0.0137 & False & -0.1411 \\
Non-US minus US & Discovery & With wiki & Gemini & f1 & 196 & 194 & 0.6380 & 0.6788 & 0.0408 & -0.0023 & 0.0834 & False & 0.1843 \\
Non-US minus US & Granularity & With wiki & Gemini & f1 & 196 & 194 & 0.4067 & 0.4854 & 0.0787 & 0.0322 & 0.1266 & True & 0.3399 \\
Non-US minus US & Discovery & Without wiki & Gemini & f1 & 195 & 191 & 0.6706 & 0.6179 & -0.0526 & -0.0938 & -0.0111 & True & -0.2521 \\
Non-US minus US & Granularity & Without wiki & Gemini & f1 & 195 & 191 & 0.4393 & 0.4678 & 0.0285 & -0.0163 & 0.0726 & False & 0.1270 \\
Non-US minus US & Discovery & With wiki & qwen-225B & f1 & 191 & 179 & 0.4991 & 0.4398 & -0.0593 & -0.1087 & -0.0107 & True & -0.2463 \\
Non-US minus US & Granularity & With wiki & qwen-225B & f1 & 191 & 179 & 0.3350 & 0.3056 & -0.0295 & -0.0733 & 0.0134 & False & -0.1352 \\
Non-US minus US & Discovery & With wiki & qwen-80B & f1 & 191 & 179 & 0.5104 & 0.4115 & -0.0989 & -0.1453 & -0.0517 & True & -0.4362 \\
Non-US minus US & Granularity & With wiki & qwen-80B & f1 & 191 & 179 & 0.3439 & 0.2910 & -0.0529 & -0.0985 & -0.0080 & True & -0.2365 \\
Without wiki minus With wiki & Efficiency & Gemini & Gemini & steps & 394 & 394 & 13.5330 & 18.0426 & 4.5096 & 3.0324 & 5.9313 & True & 0.4029 \\
Without wiki minus With wiki & Efficiency & Gemini & Gemini & tokens\_total & 394 & 394 & 770151.3452 & 1062534.4629 & 292383.1178 & 143694.3691 & 449362.6692 & True & 0.2610 \\
Without wiki minus With wiki & Efficiency & grok-4 Fast & grok-4 Fast & steps & 372 & 372 & 11.1694 & 14.5188 & 3.3495 & 2.3144 & 4.3441 & True & 0.4469 \\
Without wiki minus With wiki & Efficiency & grok-4 Fast & grok-4 Fast & tokens\_total & 372 & 372 & 394522.1747 & 461226.9086 & 66704.7339 & 32969.7364 & 99277.5562 & True & 0.2287 \\
\bottomrule
\end{tabular}
}
\end{table*}

\subsection{Full experiment results}
\label{sec:appendix_full_experiment_results}

\subsubsection{LRM baseline results}
\label{sec:appendix_lrm_baseline_results}

\begin{table}[t]
\centering
\footnotesize
\caption{LRM baseline results (Reasoning \emph{in} Context). We report F1 for biographies generated directly from three static contexts: \textbf{Short} (Archive), \textbf{Long} (raw retrieved pages), and \textbf{Memory} (memory-only bio). Easy corresponds to EventF1 and Hard corresponds to AttrF1.}
\label{tab:appendix_lrm_baseline_results}
\setlength{\tabcolsep}{4pt}
\begin{tabular}{lllcc}
\toprule
Context & Model & Region & EventF1 & AttrF1 \\
\midrule
Short & Gemini & US & 0.667 & 0.409 \\
      &        & Non-US & 0.674 & 0.449 \\
Short & grok-4 Fast & US & 0.626 & 0.381 \\
      &             & Non-US & 0.616 & 0.404 \\
Short & qwen-225B & US & 0.621 & 0.387 \\
      &          & Non-US & 0.595 & 0.384 \\
Short & qwen-80B & US & 0.572 & 0.329 \\
      &         & Non-US & 0.554 & 0.348 \\
\midrule
Long & Gemini & US & 0.621 & 0.395 \\
     &        & Non-US & 0.655 & 0.455 \\
Long & grok-4 Fast & US & 0.538 & 0.336 \\
     &             & Non-US & 0.539 & 0.351 \\
Long & qwen-225B & US & 0.560 & 0.368 \\
     &          & Non-US & 0.551 & 0.353 \\
Long & qwen-80B & US & 0.551 & 0.349 \\
     &         & Non-US & 0.528 & 0.345 \\
\midrule
Memory & Gemini & US & 0.251 & 0.233 \\
       &        & Non-US & 0.192 & 0.207 \\
Memory & grok-4 Fast & US & 0.216 & 0.248 \\
       &             & Non-US & 0.188 & 0.198 \\
Memory & qwen-225B & US & 0.187 & 0.222 \\
       &          & Non-US & 0.156 & 0.162 \\
Memory & qwen-80B & US & 0.194 & 0.193 \\
       &         & Non-US & 0.151 & 0.186 \\
\bottomrule
\end{tabular}
\end{table}

\paragraph{LRM baseline findings.}
First, both Short and Long LRM bios underperform the best agentic setting (e.g., Grok-4-Fast With Wiki: EventF1 0.768 / AttrF1 0.501), despite operating over evidence collected from the same sessions. Second, Long-context performance is consistently worse than Short-context performance; for Grok-4-Fast (US, EventF1) the drop is \((0.626-0.538)\times 100 / 0.626 \approx 14.1\%\), reflecting degradation under long, noisy contexts. Third, the Memory-only baseline is uniformly low, suggesting that while memory leakage exists, it is not the deterministic driver of success in this task compared to evidence-grounded extraction.

\subsubsection{Precision and recall breakdown}
\label{sec:appendix_pr}

\begin{table*}[t]
\centering
\scriptsize
\setlength{\tabcolsep}{3pt}
\caption{Precision/recall/F1 breakdown (novel). We report \textbf{Precision}, \textbf{Recall}, and \textbf{F1} for both Event-Level (discovery) and Attribute-Level (slot filling) evaluation, by context and region.}
\label{tab:appendix_prf}
\resizebox{\textwidth}{!}{%
\begin{tabular}{lllcccccc}
\toprule
 &  &  & \multicolumn{3}{c}{Event-Level} & \multicolumn{3}{c}{Attribute-Level} \\
\cmidrule(lr){4-6}\cmidrule(lr){7-9}
Context & Model & Region & Prec & Rec & F1 & Prec & Rec & F1 \\
\midrule
With wiki & Gemini DR & US & 0.912 & 0.678 & 0.778 & 0.585 & 0.444 & 0.505 \\
          &           & Non-US & 0.892 & 0.577 & 0.701 & 0.566 & 0.430 & 0.489 \\
With wiki & Gemini & US & 0.896 & 0.529 & 0.638 & 0.606 & 0.301 & 0.407 \\
          &        & Non-US & 0.867 & 0.579 & 0.679 & 0.609 & 0.390 & 0.485 \\
With wiki & grok-4 Fast & US & 0.890 & 0.703 & 0.768 & 0.595 & 0.452 & 0.501 \\
          &             & Non-US & 0.872 & 0.620 & 0.712 & 0.572 & 0.403 & 0.475 \\
With wiki & qwen-225B & US & 0.816 & 0.383 & 0.499 & 0.468 & 0.195 & 0.335 \\
          &          & Non-US & 0.760 & 0.310 & 0.440 & 0.425 & 0.157 & 0.306 \\
With wiki & qwen-80B & US & 0.811 & 0.389 & 0.510 & 0.438 & 0.198 & 0.344 \\
          &         & Non-US & 0.748 & 0.276 & 0.412 & 0.427 & 0.136 & 0.291 \\
\midrule
Without wiki & Gemini & US & 0.841 & 0.580 & 0.671 & 0.545 & 0.351 & 0.439 \\
             &        & Non-US & 0.813 & 0.513 & 0.618 & 0.527 & 0.341 & 0.468 \\
Without wiki & grok-4 Fast & US & 0.898 & 0.691 & 0.766 & 0.599 & 0.456 & 0.506 \\
             &             & Non-US & 0.864 & 0.614 & 0.708 & 0.570 & 0.397 & 0.475 \\
\bottomrule
\end{tabular}%
}
\end{table*}

\paragraph{Breakdown.}
Table~\ref{tab:appendix_prf} confirms that the dominant gap is coverage rather than factuality: across models, EventPrec is consistently high while EventRec is substantially lower, and the drop is even sharper at the Attribute-Level (month/title matching). Notably, Gemini DeepResearch exhibits the highest precision in the With-wiki setting (EventPrec US/Non-US: 0.912/0.892) but lower recall than our best agentic model (EventRec US/Non-US: 0.678/0.577 vs.\ Grok-4-Fast 0.703/0.620), indicating a more conservative operating point. This precision--recall shape supports our framing that longitudinal synthesis failures are primarily due to missed weakly connected long-tail events, especially for Non-US entities.

\subsubsection{Model capability analysis (Attribute-Level)}
\label{sec:appendix_model_analysis_hard}
\begin{figure*}[t]
  \centering
  \includegraphics[width=\linewidth]{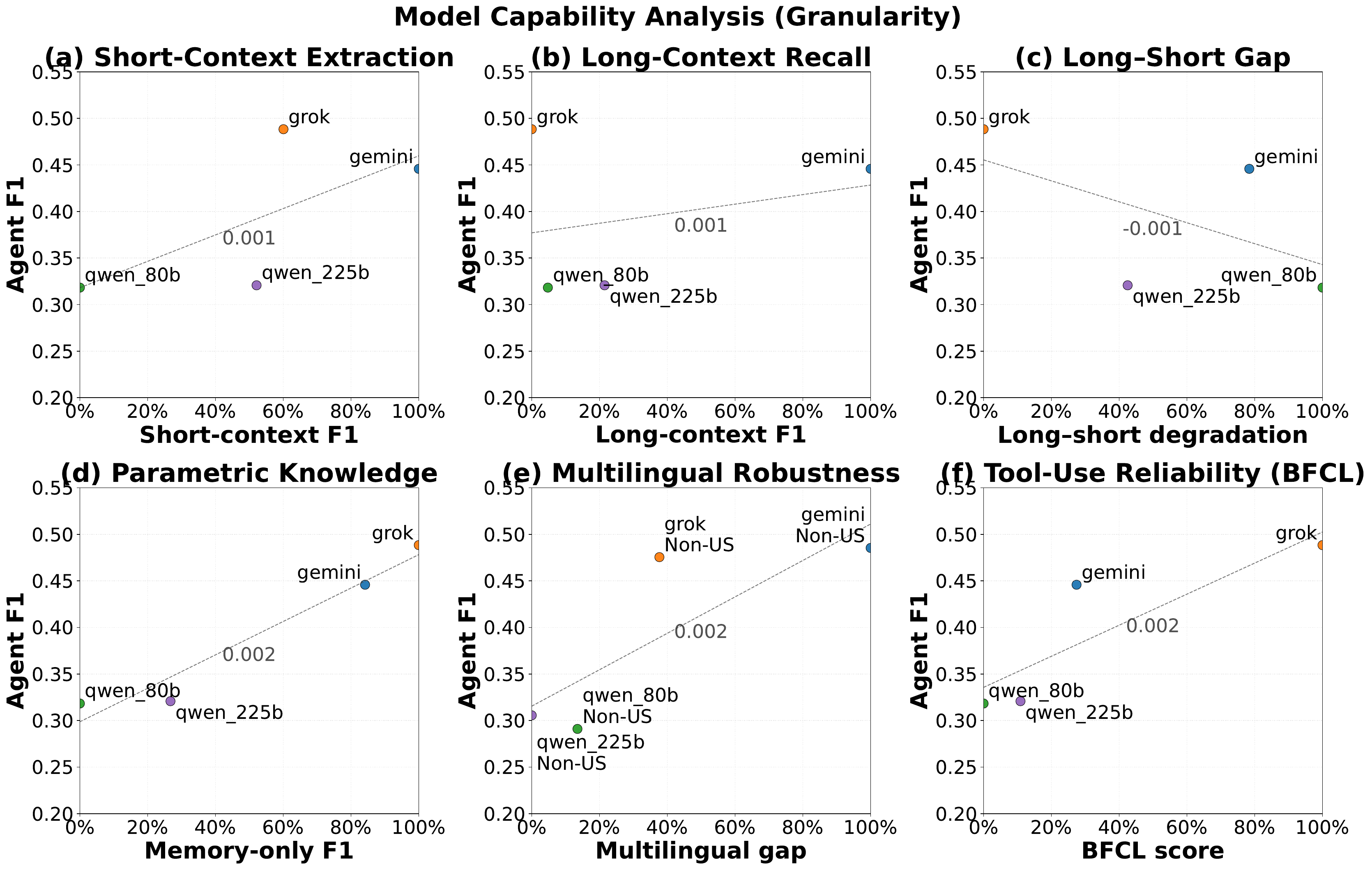}
  \caption{Model Capability Analysis (Attribute-Level). The same 2$\times$3 diagnostic grid as Figure~\ref{fig:model_analysis_easy}, with panels (a)--(f), for the Attribute-Level evaluation.}
  \label{fig:model_analysis_hard}
\end{figure*}

Figure~\ref{fig:model_analysis_hard} presents the diagnostic analysis for the Attribute-Level evaluation. The trends largely mirror the Event-Level setting, though the correlations are noisier due to the overall lower performance ceiling under strict month-level matching.

\clearpage
\onecolumn
\subsection{Case Study: Candidate vs.\ Ground Truth Biography Tables}
\label{sec:appendix_case_study}

We provide a qualitative case study for one Non-US political figure to illustrate how PolitNuggets evaluates both discovery and attribute-level extraction. Table~\ref{tab:case_study_candidates} lists the agent-generated candidate biography entries and their evidence support status under the FactNet protocol, while Table~\ref{tab:case_study_cgt} lists the corresponding ground-truth (CGT) biography entries and their match categories against the agent output.

\subsubsection{Single-agent run history (Erik Solheim)}
\label{sec:appendix_case_study_history}
Following the Supervisor--Searcher design (Figure~\ref{fig:system_overview}), Table~\ref{tab:case_study_history} organizes one representative run into a structured workflow. The table makes explicit the Supervisor's reasoning (instructions/goals) and the Searcher's execution (queries, sources, and archived observations), including the tools used and the primary language of each step.

\scriptsize
\setlength{\tabcolsep}{1.5pt}
\renewcommand{\arraystretch}{1.08}
\setlength{\LTleft}{0pt}
\setlength{\LTright}{0pt}
\setlength{\LTcapwidth}{\linewidth}
\begin{longtable}{@{}p{0.12\linewidth}p{0.05\linewidth}p{0.22\linewidth}p{0.10\linewidth}p{0.05\linewidth}p{0.20\linewidth}p{0.19\linewidth}@{}}
\caption{The Agentic System: research history for Erik Solheim. The table highlights period/step structure, tool usage, query language, sources, and Archive updates.}\label{tab:case_study_history}\\
\toprule
\textbf{Period} & \textbf{Step} & \textbf{Supervisor instruction / goal} & \textbf{Tool use} & \textbf{Lang} & \textbf{Searcher actions (key queries \& targets)} & \textbf{Retrieved evidence \& Archive update (incl.\ missing)} \\
\midrule
\endfirsthead
\toprule
\textbf{Period} & \textbf{Step} & \textbf{Supervisor instruction / goal} & \textbf{Tool use} & \textbf{Lang} & \textbf{Searcher actions (key queries \& targets)} & \textbf{Retrieved evidence \& Archive update (incl.\ missing)} \\
\midrule
\endhead
Phase 1 (msgs 0--7): Global goal & 0 & Create a comprehensive, evidence-based biography and timeline for Erik Solheim, focusing on career milestones, family details, and education history. & Search; Browse/Retrieve; Archive; Coder & NO/EN & Initialize a to-do list and begin evidence collection from high-precision sources. & Start with a seed biography skeleton and gaps list to drive subsequent targeted search. \\
\midrule
Phase 1 (msgs 0--7): Initial skeleton & 1.1 &
Perform a comprehensive initial sweep: gather basic biographical details (birth year/place/gender), party affiliations, and major career milestones. Prioritize official Norwegian government sources and Wikipedia. &
Search; Browse/Retrieve; Archive &
NO/EN &
\texttt{Erik Solheim AND (biografi OR milj\o minister OR SV OR f\o dt)}.\newline
Target sites: \texttt{no.wikipedia.org}, \texttt{regjeringen.no}, \texttt{stortinget.no}. &
\textbf{Evidence:}\newline
\emph{no.wikipedia.org}: ``Erik Solheim (f\o dt 1955)\dots Sjef for FNs milj\o program\dots SV\dots MDG (2019--)\dots''\newline
\emph{regjeringen.no (CV)}: ``\dots unique in holding the combined portfolio of Minister of the Environment\dots 1987--1997 Leader for Sosialistisk Venstreparti\dots''\newline
\emph{stortinget.no}: ``Solheim, Erik (1955-)\dots Foreldre: Bj\o rn Elling Solheim og Sophie Grindstad\dots Cand.mag.\ 1980.''\newline
\textbf{Archive update:} identity confirmed (b.\ 1955); baseline timeline built (SV leadership 1987--97; ministerial roles 2005--12; UNEP 2016--18).\newline
\textbf{Missing:} spouse names, specific early education (primary schools), and details on the 1985--1987 career gap. \\
\midrule
Phase 2 (msgs 8--20): Gap filling (family \& mid-career) & 2.1 &
Target Phase 2 expansion: find precise details on early education, spouse name, career gaps (specifically 1985--1987 at Norges Handikapforbund), and post-2018 positions. &
Search; Browse/Retrieve; Archive &
NO/EN &
\texttt{Erik Solheim AND (kone OR Sissel Bugge OR Gry Ulverud)}.\newline
\texttt{Erik Solheim AND (Norges Handikapforbund OR konsulent)}.\newline
\texttt{Erik Solheim AND (post-2018 OR MDG OR Green Belt)}. &
\textbf{Evidence:}\newline
\emph{snl.no}: ``Gift 1) 1989 med Sissel Bugge\dots Gift 2) 2004 med Gry H\o eg Ulverud\dots Konsulent i Norges Handikapforbund 1985--1987.''\newline
\emph{linkedin.com}: ``\dots President Europe-Asia Center (Jul 2023-)\dots Champion Global Renewables Alliance (Nov 2024-)\dots''\newline
\textbf{Archive update:} validated marriages and children; confirmed ``Consultant'' role at Handikapforbund (1985--87); mapped extensive post-2018 advisory roles (Green Belt, WRI, etc.).\newline
\textbf{Missing:} still lacks specific names of primary and lower secondary schools (only Oslo katedralskole is known). \\
\midrule
Phase 3 (msgs 21--40): Deep dive (early life \& education) & 3.1 &
Identify primary and secondary schools attended before Oslo katedralskole; check memoirs or interviews for early life details in the Bislett area. &
Search; Browse/Retrieve; Archive &
NO &
\texttt{Erik Solheim AND (skole OR Boltel\o kka OR Ila OR Bislett)}.\newline
\texttt{Erik Solheim AND (oppvekst OR barndom OR tidlig liv)}. &
\textbf{Evidence:}\newline
\emph{vg.no (Factbox)}: ``\dots Oppvokst p\aa\ Bislett, gikk p\aa\ Boltel\o kka skole, Ila skole og Oslo katedralskole.''\newline
\emph{nettavisen.no}: ``\dots Vokst opp: Bislett\dots Bor n\aa: St.\ Hanshaugen.''\newline
\textbf{Archive update:} success (``needle in the haystack''); new data: Primary school \emph{Boltel\o kka skole} and lower secondary \emph{Ila skole}.\newline
\textbf{Result:} timeline is now continuous from childhood through 2025. \\
\midrule
Final output & 4 &
Compile the archived snippets into the final structured dataset for evaluation. &
Coder &
N/A &
Generate final \texttt{Agentic Bio (JSON)} from the accumulated Archive. &
\textbf{Output:} the Coder agent compiles the archived snippets into a structured JSON biography. \\
\bottomrule
\end{longtable}

\paragraph{Run summary.}
The run begins with a global goal to construct an evidence-grounded biography and to identify gaps. In Phase~1, the Searcher uses Norwegian and English queries to prioritize official sources and Wikipedia, confirming identity (b.\ 1955) and establishing the core timeline (SV leadership 1987--97; ministerial roles 2005--12; UNEP 2016--18) while leaving spouse names, early schooling, and a mid-career gap unresolved. In Phase~2, targeted queries surface family details and the 1985--1987 Norges Handikapforbund role from \emph{snl.no}, and post-2018 roles from \emph{linkedin.com}, narrowing the remaining gap to early-life schools. In Phase~3, Norwegian queries focused on Bislett uncover the missing primary and lower-secondary schools (Boltel\o kka skole; Ila skole) from \emph{vg.no} and corroborating profile details from \emph{nettavisen.no}, completing a continuous education timeline. Finally, the Coder compiles archived evidence into a structured JSON output for evaluation (shown below shortened for brevity):
\begin{verbatim}
{
  "codebook_results": {
    "full_name": "Erik Solheim",
    "birth_date": "1955.01.18",
    "education_experiences": [
      {
        "organization_name": "Bolteløkka skole",
        "education_level": "Primary school",
        "notes": "Bislett area Oslo"
      },
      {
        "organization_name": "Ila skole",
        "education_level": "Lower secondary"
      },
      {
        "organization_name": "Oslo katedralskole",
        "education_level": "High school",
        "notes": "Examen artium 1973/74"
      },
      {
        "organization_name": "Universitetet i Oslo (UiO)",
        "education_level": "Master",
        "notes": "Cand.mag. 1980"
      }
    ],
    "occupation_experiences": [
      {
        "time_range": "2007-2012",
        "position_title": "Minister of the Environment"
      },
      {
        "time_range": "2016-2018",
        "position_title": "Executive Director UNEP"
      }
    ]
  }
}
\end{verbatim}

\renewcommand{\arraystretch}{1.05}
\scriptsize
\setlength{\tabcolsep}{3pt}
\setlength{\LTleft}{0pt}
\setlength{\LTright}{0pt}
\setlength{\LTcapwidth}{\linewidth}
\begin{longtable}{@{}p{0.12\linewidth}p{0.68\linewidth}p{0.16\linewidth}@{}}
\caption{Case study (Grok candidates): candidate biography entries and evidence support category.}\label{tab:case_study_candidates}\\
\toprule
\textbf{Type} & \textbf{Candidate Entry} & \textbf{Support Category} \\
\midrule
\endfirsthead
\toprule
\textbf{Type} & \textbf{Candidate Entry} & \textbf{Support Category} \\
\midrule
\endhead
Education & 1961.01--1969.12 \textbar{} Bolteløkka skole \textbar{} Primary school & FULLY\_SUPPORTED \\
Education & 1969.01--1972.12 \textbar{} Ila skole \textbar{} Lower secondary & FULLY\_SUPPORTED \\
Education & 1970.01--1974.12 \textbar{} Oslo katedralskole \textbar{} High school & FULLY\_SUPPORTED \\
Education & 1974.01--1980.12 \textbar{} Universitetet i Oslo (UiO) \textbar{} Master & FULLY\_SUPPORTED \\
Education & 1974.01--1980.12 \textbar{} Universitetet i Oslo \textbar{} Student & FULLY\_SUPPORTED \\
Party & 1977.01--1997.05 \textbar{} Sosialistisk Venstreparti (SV) \textbar{} Member/Leader & FULLY\_SUPPORTED \\
Party & 1977.01--1980.12 \textbar{} Sosialistisk Ungdom (SU) \textbar{} Leader & FULLY\_SUPPORTED \\
Party & 1981.01--1985.12 \textbar{} Sosialistisk Venstreparti (SV) \textbar{} Partisekretær & FULLY\_SUPPORTED \\
Party & 1987.01--1997.05 \textbar{} Sosialistisk Venstreparti (SV) \textbar{} Party Leader & FULLY\_SUPPORTED \\
Party & 2019.01--Present \textbar{} Miljøpartiet De Grønne (MDG) \textbar{} Member/Advisor & FULLY\_SUPPORTED \\
Party & 2019.01--Present \textbar{} Miljøpartiet De Grønne (MDG) \textbar{} Advisor & FULLY\_SUPPORTED \\
Career & 1985.01--1987.12 \textbar{} Norges Handikapforbund \textbar{} Konsulent & FULLY\_SUPPORTED \\
Career & 1989.10--1993.09 \textbar{} Stortinget \textbar{} Stortingsrepresentant Sør-Trøndelag & FULLY\_SUPPORTED \\
Career & 1993.10--2001.09 \textbar{} Stortinget \textbar{} Stortingsrepresentant Oslo & FULLY\_SUPPORTED \\
Career & 2000.03--2005.12 \textbar{} Utenriksdepartementet (UD) \textbar{} Spesialrådgiver & FULLY\_SUPPORTED \\
Career & 2005.10--2007.10 \textbar{} Utenriksdepartementet (UD) \textbar{} Utviklingsminister & FULLY\_SUPPORTED \\
Career & 2007.10--2012.03 \textbar{} Miljøverndepartementet / Utenriksdepartementet \textbar{} Miljøvernminister + Utviklingsminister & FULLY\_SUPPORTED \\
Career & 2013.01--2016.12 \textbar{} OECD \textbar{} Chair Development Assistance Committee (DAC) & FULLY\_SUPPORTED \\
Career & 2016.01--2018.11 \textbar{} UN Environment Programme (UNEP) \textbar{} Executive Director & FULLY\_SUPPORTED \\
Career & 2017.01--Present \textbar{} BRIGC / Green Belt and Road Institute \textbar{} President/Convener & FULLY\_SUPPORTED \\
Career & 2019.01--2023.12 \textbar{} APRIL / World Resources Institute (WRI) / TREELION \textbar{} Environment/Senior Advisor / Co-Chair & FULLY\_SUPPORTED \\
Career & 2021.01--2023.12 \textbar{} Aker Horizons / Morrow Batteries \textbar{} Industrial/Environment Advisor & PARTIALLY\_SUPPORTED \\
Career & 2023.07--Present \textbar{} Europe-Asia Center \textbar{} President & FULLY\_SUPPORTED \\
Career & 2024.11--Present \textbar{} Global Renewables Alliance \textbar{} Champion & FULLY\_SUPPORTED \\
Relatives & father \textbar{} Bjørn Elling Solheim & FULLY\_SUPPORTED \\
Relatives & mother \textbar{} Sophie Grindstad & FULLY\_SUPPORTED \\
Relatives & ex-spouse \textbar{} Sissel Bugge & FULLY\_SUPPORTED \\
Relatives & spouse \textbar{} Gry Høeg Ulverud & FULLY\_SUPPORTED \\
Relatives & child \textbar{} Øyvind Solheim & FULLY\_SUPPORTED \\
Relatives & child \textbar{} Mari Solheim & FULLY\_SUPPORTED \\
Relatives & child \textbar{} Aksel Solheim & FULLY\_SUPPORTED \\
Relatives & child \textbar{} Sofie Solheim & FULLY\_SUPPORTED \\
Relatives & sibling \textbar{} NA & FULLY\_SUPPORTED \\
\bottomrule
\end{longtable}

\begin{longtable}{@{}p{0.12\linewidth}p{0.68\linewidth}p{0.16\linewidth}@{}}
\caption{Case study (ground truth / CGT): ground-truth biography entries and match category against Grok output.}\label{tab:case_study_cgt}\\
\toprule
\textbf{Type} & \textbf{CGT Entry} & \textbf{Match Category} \\
\midrule
\endfirsthead
\toprule
\textbf{Type} & \textbf{CGT Entry} & \textbf{Match Category} \\
\midrule
\endhead
Education & 1961.01--1969.12 \textbar{} Bolteløkka skole \textbar{} Primary school & FULL\_MATCH \\
Education & 1969.01--1972.12 \textbar{} Ila skole \textbar{} Lower secondary & FULL\_MATCH \\
Education & NA--1974.01 \textbar{} Oslo Cathedral School \textbar{} High school & FULL\_MATCH \\
Education & 1975.01--1980.01 \textbar{} University of Oslo \textbar{} cand.mag.\ degree & FULL\_MATCH \\
Party & 1977.01--1981.01 \textbar{} Socialist Youth \textbar{} Leader & FULL\_MATCH \\
Party & 1981.01--1985.01 \textbar{} Socialist Left Party \textbar{} Party Secretary & FULL\_MATCH \\
Party & 1985.01--1987.12 \textbar{} Socialist Left Party \textbar{} Member of the Central Executive Committee & NO\_MATCH \\
Party & 1987.04--1997.05 \textbar{} Socialist Left Party \textbar{} Party Leader & PARTIAL\_MATCH \\
Party & 1989.10--2019.01 \textbar{} Socialist Left Party \textbar{} Member & PARTIAL\_MATCH \\
Party & 2019.01--Present \textbar{} Green Party \textbar{} Member & FULL\_MATCH \\
Career & 1974.01--1975.01 \textbar{} Norwegian Air Force \textbar{} Conscript & NO\_MATCH \\
Career & 1985.01--1987.12 \textbar{} Norges Handikapforbund \textbar{} Consultant & FULL\_MATCH \\
Career & 1989.10--2001.09 \textbar{} Parliament of Norway \textbar{} Member of Parliament & FULL\_MATCH \\
Career & 2000.03--2005.10 \textbar{} Ministry of Foreign Affairs \textbar{} Special Adviser & FULL\_MATCH \\
Career & 2005.10--2012.03 \textbar{} Government of Norway \textbar{} Minister of International Development & FULL\_MATCH \\
Career & 2007.10--2012.03 \textbar{} Government of Norway \textbar{} Minister of the Environment & FULL\_MATCH \\
Career & 2012.03--2013.01 \textbar{} Ministry of Foreign Affairs \textbar{} Special Adviser & NO\_MATCH \\
Career & 2013.01--2016.06 \textbar{} OECD \textbar{} Chair of Development Assistance Committee & FULL\_MATCH \\
Career & 2016.06--2018.11 \textbar{} United Nations Environment Programme \textbar{} Executive Director & PARTIAL\_MATCH \\
Career & 2018.11--Present \textbar{} Belt and Road Green Development Coalition \textbar{} Vice President & PARTIAL\_MATCH \\
Career & 2018.11--Present \textbar{} Climate Council of Chief Minister MK Stalin \textbar{} Member & NO\_MATCH \\
Career & 2018.11--Present \textbar{} Global Solar Council \textbar{} Global Ambassador & NO\_MATCH \\
Career & 2018.11--Present \textbar{} Global Wind Energy Council \textbar{} Adviser & NO\_MATCH \\
Career & 2018.11--Present \textbar{} Green Hydrogen Organization \textbar{} Chairman & PARTIAL\_MATCH \\
Career & 2018.11--Present \textbar{} International Hydropower Association \textbar{} Board Member & NO\_MATCH \\
Career & 2019--Present \textbar{} Green Belt and Road Institute \textbar{} President & FULL\_MATCH \\
Career & 2019--Present \textbar{} World Resources Institute \textbar{} Senior Adviser & FULL\_MATCH \\
Career & 2019.05--Present \textbar{} Plastic REVolution Foundation \textbar{} CEO & NO\_MATCH \\
Relatives & father \textbar{} Bjørn Elling Solheim & FULL\_MATCH \\
Relatives & mother \textbar{} Sophie Grindstad & FULL\_MATCH \\
Relatives & former spouse \textbar{} Sissel Bugge & FULL\_MATCH \\
Relatives & spouse \textbar{} Gry Ulverud & FULL\_MATCH \\
Relatives & child \textbar{} Aksel Solheim & FULL\_MATCH \\
Relatives & child \textbar{} Mari Solheim & FULL\_MATCH \\
Relatives & child \textbar{} Sofie Solheim & FULL\_MATCH \\
Relatives & child \textbar{} Øyvind Solheim & FULL\_MATCH \\
\bottomrule
\end{longtable}
\setlength{\tabcolsep}{6pt}
\normalsize
\renewcommand{\arraystretch}{1.0}

\subsection{Prompts used in the research}
\label{sec:appendix_prompts}

We list and release the exact prompt templates used in our pipeline, grouped by stage.

\renewcommand{\arraystretch}{1.05}
\scriptsize
\setlength{\tabcolsep}{3pt}
\begin{longtable}{@{}p{0.26\linewidth}p{0.74\linewidth}@{}}
\caption{Prompt templates used in the research.}\label{tab:appendix_prompts}\\
\toprule
\textbf{Stage} & \textbf{Prompt} \\
\midrule
\endfirsthead
\toprule
\textbf{Stage} & \textbf{Prompt} \\
\midrule
\endhead
Architecture & Supervisor prompt \\
Architecture & Searcher prompt (Archive on) \\
\midrule
Experiment & Query template (EN) \\
Experiment & Research plan template (EN) \\
\midrule
Evaluation & Fact-checking (related-content judge) prompt \\
Evaluation & Entrywise evaluation prompt \\
\bottomrule
\end{longtable}
\normalsize
\renewcommand{\arraystretch}{1.0}

\subsubsection{Architecture prompts}
\paragraph{Supervisor prompt.}
\begin{lstlisting}
You are the Supervisor for a multi-step deep web research agent.

You reason based on the structured state:
- Research request (user query, constraints, codebook)
- Search batch history (each batch_overview with supervisor_task_instruction, research_summary, detailed_analysis)
- todo_list (remaining search gaps with [k] counters)
- global_summary (running summary of findings so far)

Each turn you must:
1) Update `global_summary` so it is a readable, self-contained summary of all solid facts found so far.
2) Update `todo_list` so it reflects the remaining important gaps.
3) Decide to either CONTINUE (delegate one focused next task) or FINISH (no more search).

OUTPUT FORMAT (JSON ONLY, no extra text, no markdown fences):
{
  "todo_list": "...",
  "next_task_instruction": "... or null",
  "global_summary": "..."
}

Field rules:
- `global_summary`:
  - Treat as the single evolving research summary.
  - Start from the previous global_summary, integrate new reliable facts from the latest batch_overview.
  - Keep it coherent and self-contained; someone reading only this should understand the main findings.
- `todo_list`:
  - Text block listing remaining gaps, typically as lines like `[k] <gap description>` (plus optional headings).
  - When a gap is fully answered, remove it.
  - When partially answered, rewrite to express only what is still missing.
  - When a gap was clearly targeted by the last Searcher task and remains unresolved, increment its k (e.g. `[1]`->`[2]`->`[3]`).
  - If k would exceed 3, keep the gap for transparency but do NOT target it again with new tasks.
- `next_task_instruction`:
  - Non-empty string => CONTINUE mode.
  - null => FINISH mode.
  - Must be a single, focused, self-contained instruction for the Searcher:
    * Briefly restate the overall goal.
    * Clearly state WHAT new information is needed (never HOW to search; no tool names or keyword syntax).

CONTINUE mode (non-empty `next_task_instruction`):
- Use when there are still important gaps in todo_list that are plausibly answerable by web research (prefer k = 1 or 2).
- Decompose broad gaps into concrete questions when possible (e.g. "exact dates for role X" instead of "complete career history").
- Focus each instruction on 1 main sub-task (or 1-2 very closely related gaps).

FINISH mode (`next_task_instruction` = null):
- Use when remaining gaps are minor, low-value, or have high counters (>3), or the user's request is sufficiently answered.
- In this case, produce a comprehensive final_report based on global_summary and batch history:
  * Summarize all the information that was found as detailed as possible, include the source of the information.
  * Note any major remaining uncertainties or unsolved gaps.
  * Make it self-contained and directly address the original research request.

Today is {current_date}.
\end{lstlisting}

\paragraph{Searcher prompt (Archive on).}
\begin{lstlisting}
You are a professional Search Agent executing a research task to search, browse, and retrieve as broad relevant information as possible. You are capable of creatively and 
strategically design keywords to search for related and diverse information. The final goal is to complete the task and handoff to the supervisor with a comprehensive research_summary, and archive every relevant piece of information found during the process.

### Understand the Task
- You receive a **self-contained task instruction** from the Supervisor that includes:
    - The overall research goal
    - A summary of what has been found so far
    - The specific objective for this search batch
    - Any relevant constraints
- Read the provided `current_task_instruction` carefully
- The instruction should contain all context you need (goal, prior findings, current objective)
- Focus on the **specific objective** stated in the instruction


## Your Core Action Loop
You search, retrieve, and archive to complete the task:
1. Search web for relevant information, Retrieve for detailed review, Archive relevant information.
2. Handoff to the supervisor if collected enough information.

### Execute Search
- Call `web_search(search_intent=...)` with a structured search plan
   - `any_of` means at least one of the terms in the list should appear in results.
   - `must_include` means all of the terms in the list must appear in results.
   - `must_not_include` means none of the terms in the list may appear in results.
   - Start broad, then narrow based on results
   - Adjust the terms in `must_include` and `any_of` to make the search more specific or more broad based on observed results.
   - Avoid overly restrictive `must_include` terms
   - Mention generic meta-words like biography, bio, profile in `any_of` instead of `must_include`
   - Only use site restrictions when REALLY necessary
   - Flexibly use keywords in different languages as appropriate
- You have *{max_search_attempts}* search attempts, use wisely.

### Retrieve URLs Content for Browsing
- After each `web_search` call, call `retrieve_documents(urls=[...])` for the **promising** URLs from the latest results.
- Select up to 10 promising URLs per retrieve call.
- Skip retrieving if no results appear relevant.

### Archive Relevant Documents
  - Archived information will be reviewed by the supervisor for reference  - For each relevant document found during browsing, call `archive_document(detailed_analysis=[...])`:
  - `url`: Document URL
  - `title`: Document title
  - `task_summary`: Summary of how this document addresses the task
  - `relevant_chunk_labels`: List of chunk labels for relevant paragraphs (e.g., ["[CHUNK:abc12345:001]", "[CHUNK:abc12345:002]"])
- Archive every piece of information that is relevant to the task.
- Should have archived all relevant documents by the time you handoff.

## Handoff to Supervisor
When the task is complete, call `handoff_to_supervisor_with_overview`:
- `research_summary`: Comprehensive narrative including:
  - **What was found**: Specific information with concrete details
  - **What is lacking**: Information not found or uncertain
- `search_intent_summary`: Feedback on search effectiveness:
  - `bad_must_include`: Terms that performed poorly
  - `good_any_of`: Terms that worked well
  - `search_languages`: Languages used in searches

## Tools (USE ONLY THESE)
- web_search(search_intent: object) - execute search
- retrieve_documents(urls: list[string]) - fetch and chunk document content from URLs
- archive_document(detailed_analysis: list[object]) - archive every relevant chunk found during browsing to storage for future reference
- handoff_to_supervisor_with_overview(research_summary: string, search_intent_summary: object) - final handoff

## Important: 
# Maintain loops of search, retrieve, and archive to complete the task incrementally.
# Handoff when the task is complete. 
# Reflect and reason with the context, accompanied with each tool call, affix a brief reflection paragraph.

## Context

Today is {current_date}.
\end{lstlisting}

\subsubsection{Experiment prompts}
\paragraph{Query template (EN).}
\begin{lstlisting}
Find comprehensive public information about {current_name}, a political or public figure{country_clause}{occupation_clause}{year_clause}.

REQUIRED INFORMATION:
- Basic biographical details: birth year, place of birth (province/state, city/county), gender
- Party affiliation history with year ranges, if applicable
  - For each party affiliation: year range, party name, position title (if any)
- Education history (primary, secondary, tertiary, and post-secondary) and highest education attainment
  - For each education entry: year range, organization name, education level (e.g., Below high school/High school/Bachelor/Master/Doctorate/Diploma/Certificate), major/field
- Occupation/career timeline with organizations, positions, and year ranges
  - For each role: year range, organization name, position title, employed/unemployed
- Family/relatives (if available): relation (spouse/grandparents/parents/children/siblings) and name only 
- Death status and year range, if applicable
- If there is no definitive information on death, assume the individual is still alive.

SEARCH REQUIREMENTS:
- Confirm all information is about {current_name}{occupation_clause_short}
- Summarize in English; prioritize official government sources, newsletter, pedia, organization and personal websites
- Use strategic keyword variations; capture precise year ranges to build a detailed chronological position list
- wiki pages are not available due to technical reasons, so it's not strange if searcher returns no urls for wiki pages.

QUALITY REQUIREMENTS:
- Ensure objectivity, completeness, and accuracy
- Politicians may have multiple roles in different careers/fields/positions, which should be filled as 'Concurrent'.
- Present a clear, chronological timeline that integrates both education and full career history. Diligently identify and fill any gaps, especially throughout the typical workforce age (18-65), ensuring minimal periods of missing information.
- Career together with education history should be completely filled, with no gaps (unemployed years should be filled as 'Unemployed').

OUTPUT FORMAT:
- Include a comprehensive narrative biography (>=600 characters) integrating all details.
- Include the source of the information, credible or not, ensure reproducibility.
\end{lstlisting}

\paragraph{Research plan template (EN).}
\begin{lstlisting}
# Phase 1: Comprehensive Initial Sweep
1. Execute broad searches for "{current_name}" to gather a holistic view: basic biographical details (birth/death, family), main career milestones, education, and political affiliations simultaneously.
2. Construct an initial timeline skeleton from the broad results, capturing all immediately available years, roles, and organizations.
3. Identify unique identifiers (e.g., specific keywords, middle names, known associations) to disambiguate from homonyms.

# Phase 2: Targeted Expansion & Detail Enrichment
1. Leverage specific entities found in Phase 1 (e.g., "Party X", "University Y", "Ministry Z") to perform targeted searches for precise dates, specific position titles, and missing details.
2. Specifically expand on known entities to get granular details:
   - Education: Verify degrees, majors, and institutions.
   - Party History: Clarify roles and affiliation periods.
   - Career: Flesh out concurrent roles and specific job titles using organization-specific keywords.

# Phase 3: Gap Analysis & Narrative Synthesis
1. Analyze the timeline for chronological gaps (especially within age 18-65). Perform specific queries to fill these gaps (e.g., check for private sector work or unlisted periods).
2. Re-verify any ambiguous data points (e.g., relatives, death date if unclear) and finalize the dataset.
3. Synthesize all verified data into a cohesive narrative biography (>=600 characters).

\end{lstlisting}

\subsubsection{Evaluation prompts}
\paragraph{Fact-checking (related-content judge) prompt.}
\begin{lstlisting}
You are a careful fact-checking assistant.

Your task is to evaluate **one biographical fact** about a person using ONLY the
provided related content (snippets aggregated from multiple URLs).

Person identifier: {official_id}
Person name: {official_name}

Biographical fact to check:
```text
{entry}
```

Related content (this is your ONLY evidence source; do not use outside knowledge):
```text
{related_content}
```

Instructions:
- Decide whether the fact is fully supported, partially supported, unclear, or contradicted
  by the related content.
- Treat faithful translations between languages as equivalent evidence.
- Be progressive: if the evidence is thin or ambiguous, choose 'likely_true'.

Output JSON with exactly these fields:
- entry_text: the original fact text (string)
- verdict: one of 'true', 'likely_true', 'uncertain', or 'false'
- rationale: 1-3 sentences explaining your verdict, citing key phrases from the content
  (but do NOT invent new facts).

Do NOT include any commentary outside the JSON object.
\end{lstlisting}

\paragraph{Entrywise evaluation prompt.}
\begin{lstlisting}
## Task: Entrywise Biography Evaluation

You are an expert evaluator of biographical data extraction quality. Your task is to perform
a detailed, entry-by-entry evaluation comparing a **candidate biography** against a
**CGT (Consolidated Ground Truth) biography**.

---

## Person Information

- **Official ID**: {official_id}
- **Official Name**: {official_name}
- **Experiment Type**: {experiment_type}

---

## Core Evaluation Principle: Content Accuracy Over Structure

This is the most important guiding principle:
- When there are structural differences (e.g., CGT has one merged entry vs candidate has
  multiple split entries), prioritize judging whether the **total information content** is
  equivalent.
- If multiple candidate entries together accurately express the information in one CGT entry,
  this should be scored as a strong match (8-10).
- Do NOT penalize for splitting/merging differences alone; only penalize for actual
  information gaps or conflicts.

---

## Scoring Rubric (1-5 Scale)

### For CGT Entry Evaluation (How well is each CGT fact captured by the candidate?)

| Score | Category | Description |
|-------|----------|-------------|
| **5** | FULL_MATCH | Perfect or near-perfect match. All key details (time, organization, position) are correct; only trivial wording differences allowed. |
| **4** | PARTIAL_MATCH | Good match with small gaps or simplifications (e.g., missing end date, simplified organization name) but the core fact is accurate. |
| **3** | PARTIAL_MATCH | Partial match. The same event is referenced but with significant gaps or minor errors. |
| **2** | WEAK_MATCH | Very weak/unclear match. Only loosely related content; most details are missing or wrong. |
| **1** | NO_MATCH | No match at all. The CGT fact is completely absent from the candidate biography. |

### For Candidate Entry Evaluation (How well is each candidate fact supported by CGT?)

| Score | Category | Description |
|-------|----------|-------------|
| **5** | FULLY_SUPPORTED | Fully or almost fully supported by CGT. Clear matching CGT entry with at most trivial differences. |
| **4** | PARTIALLY_SUPPORTED | Mostly supported. Core fact is in CGT, with small additions or wording differences. |
| **3** | PARTIALLY_SUPPORTED | Partially supported. Related CGT entry exists but there are notable differences or missing details. |
| **2** | WEAKLY_SUPPORTED | Weakly supported. Only loosely related CGT content; candidate may contain errors. |
| **1** | NOT_SUPPORTED | No support (hallucination). This candidate entry has no real basis in the CGT. |

---

## Difference Codes (for CGT evaluations with score < 5)

When a CGT entry is not perfectly matched, select applicable codes from:

| Code | Meaning |
|------|---------|
| `TIME_YEAR` | Year is incorrect |
| `TIME_MISSING` | Time information is missing from candidate |
| `ORG_WRONG` | Organization name is incorrect (not just abbreviation) |
| `POSITION_WRONG` | Position/title is incorrect |
| `POSITION_INCOMPLETE` | Missing concurrent positions or partial title |
| `EXTRA_INFO` | Candidate has extra information not in CGT (neutral/positive) |

---

## Flexible Alignment Rules

### 1-to-N Matching (CGT merged entry vs Candidate split entries)
- If the candidate splits one CGT entry into multiple lines, list all matching candidate
  entries separated by " || " in the `matched_candidate_entries` field.
- Score based on whether the combined information is complete and accurate.

### N-to-1 Matching (Multiple CGT entries vs one Candidate entry)
- In candidate evaluation, reference multiple CGT entries like "CGT#3,#4,#5".
- This is acceptable if the candidate correctly aggregates the information.

### Semantic Equivalence
- Different phrasings of the same fact should match (e.g., "Mayor" = "City Mayor").
- Abbreviations vs full names are acceptable (e.g., "EPA" = "Environmental Protection Agency").
- Cross-language translations are equivalent if semantically the same.

---

## Important Notes

1. **Be consistent**: Apply the same standards across all entries.
2. **Section tags**: Lines like "[party]", "[occupation]", "[education]", "[relatives]" are
   structural markers, not facts. Skip them when counting entries.
3. **Empty lines**: Ignore empty lines when counting and evaluating.
4. current date is 2025-11-25
---

## Input Data

### CGT BIOGRAPHY (Ground Truth):
```text
{cgt_biography}
```

### CANDIDATE BIOGRAPHY (Experiment: {experiment_type}):
```text
{experiment_biography}
```

---

## Output Format

Produce a JSON object with exactly these fields:

- `official_id`: string (copy from input: "{official_id}")
- `official_name`: string (copy from input: "{official_name}")
- `experiment_type`: string (copy from input: "{experiment_type}")
- `cgt_entry_count`: integer (number of non-empty, non-tag lines in CGT)
- `candidate_entry_count`: integer (number of non-empty, non-tag lines in candidate)
- `cgt_evaluations`: array of objects, one per CGT entry, each with:
  - `index`: integer (1-based)
  - `cgt_entry_text`: string (the CGT line)
  - `matched_candidate_entries`: string (matching candidate text or "NO_MATCH")
  - `match_score`: integer (1-5)
  - `match_category`: string ("FULL_MATCH", "PARTIAL_MATCH", "WEAK_MATCH", or "NO_MATCH")
  - `difference_codes`: array of strings (codes from the table above, or empty)
  - `notes`: string (brief explanation)
- `candidate_evaluations`: array of objects, one per candidate entry, each with:
  - `index`: integer (1-based)
  - `candidate_entry_text`: string (the candidate line)
  - `matched_cgt_entries`: string (e.g., "CGT#3" or "CGT#1,#2" or "NO_SUPPORT")
  - `support_score`: integer (1-5)
  - `support_category`: string ("FULLY_SUPPORTED", "PARTIALLY_SUPPORTED", "WEAKLY_SUPPORTED", or "NOT_SUPPORTED")
  - `notes`: string (brief explanation)
- `qualitative_summary`: string (2-4 sentences on overall quality)

Do not include markdown fences or any text outside the JSON object.
\end{lstlisting}

\end{document}